%% file: main.tex
\input{_constants}

\arxiv % \review OR \arxiv OR \cameraready

\pdfoutput=1
\PassOptionsToPackage{prologue,dvipsnames}{xcolor}
\documentclass[10pt,twocolumn,letterpaper]{article}
\usepackage{indentfirst}
\usepackage{float}
\usepackage{bbding}
\usepackage{amssymb}
\usepackage{bm}
\usepackage{tikz}
\usepackage{colortbl}
\usepackage{tabu}
\usepackage{array}
\usepackage[accsupp]{axessibility} % Improves PDF readability for those with visual impairments.
\usetikzlibrary{spy}
\input{cvpr_header}
\begin{document}

\input{symbols_format.tex}

\title{RoGSplat: Learning Robust Generalizable Human Gaussian Splatting \\ from Sparse Multi-View Images}

\author{
    Junjin Xiao$^1$\quad  
    Qing Zhang$^{1,4*}$\quad  
    Yonewei Nie$^{2}$ \quad
    Lei Zhu$^{3}$ \quad
    Wei-Shi Zheng$^{1,4}$ \\
   \small $^1$ School of Computer Science and Engineering, Sun Yat-sen University, China \\
   \small $^2$South China University of Technology\quad
   $^3$Hong Kong University of Science and Technology (Guangzhou)\\
    \small $^4$Key Laboratory of Machine Intelligence and Advanced Computing, Ministry of Education, China\\
    %{
    % \tt\small 
    % xiaojj37@mail2.sysu.edu.cn\quad 
    % zhangq93@mail.sysu.edu.cn\\
    % \tt\small
    % zhaxu@adobe.com\quad 
    % wszheng@ieee.org
    %}
}

% *********************************
% -------------TEASER--------------
% *********************************
\twocolumn[{
\renewcommand\twocolumn[1][]{#1}%
\maketitle
\vspace{-8mm}
\begin{figure}[H]
\hsize=\textwidth %
\centering
\includegraphics[width=1\textwidth]{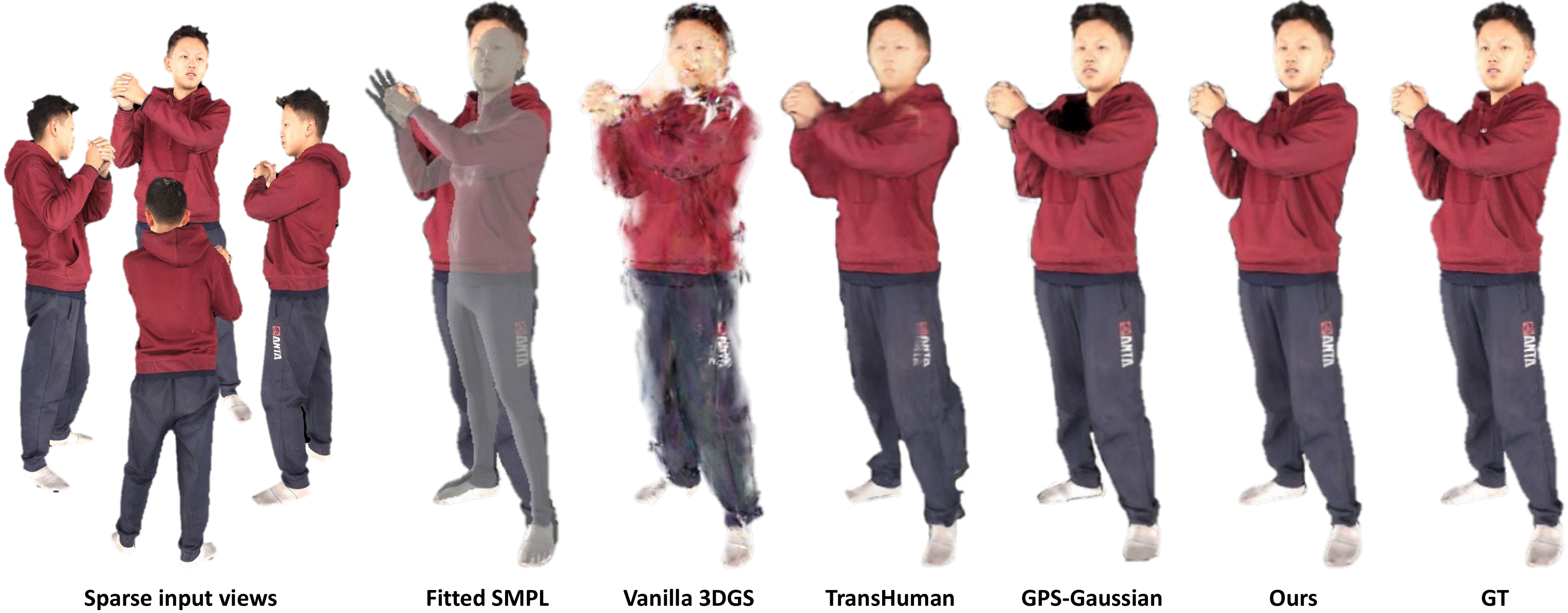}
\vspace{-6mm}
\caption{\qing{\textbf{High-fidelity human novel view synthesis.} Given very sparse-view input images (\eg, 4 views) that do not enable accurate human template estimation due to the limited overlappings, our method can robustly synthesize high-fidelity novel views in a generalizable manner, without requiring any further fine-tuning or subject-specific optimization. Compared to both NeRF-based method, \eg, TransHuman~\cite{Pan_2023_ICCV}, and 3D Gaussian Splatting (3DGS) based methods, \eg, vanilla 3DGS~\cite{3DGS} and GPS-Gaussian~\cite{zheng2024gpsgaussian}, our approach produces better result.}}
\label{fig:teaser}
\end{figure}
}]

\setlength{\parindent}{1pc}

\renewcommand{\thefootnote}{}
\footnotetext{$^*$Corresponding author (zhangq93@mail.sysu.edu.cn).}

% *********************************
% ------------ABSTRACT-------------
% *********************************
\begin{abstract}
This paper presents RoGSplat, a novel approach for synthesizing high-fidelity novel views of unseen human from sparse multi-view images, while requiring no cumbersome per-subject optimization. Unlike previous methods that typically struggle with sparse views with few overlappings and are less effective in reconstructing complex human geometry, the proposed method enables robust reconstruction in such challenging conditions. Our key idea is to lift SMPL vertices to dense and reliable 3D prior points representing accurate human body geometry, and then regress human Gaussian parameters based on the points. To account for possible misalignment between SMPL model and images, we propose to predict image-aligned 3D prior points by leveraging both pixel-level features and voxel-level features, from which we regress the coarse Gaussians. To enhance the ability to capture high-frequency details, we further render depth maps from the coarse 3D Gaussians to help regress fine-grained pixel-wise Gaussians. Experiments on several benchmark datasets demonstrate that our method outperforms state-of-the-art methods in novel view synthesis and cross-dataset generalization. Our code is available at \url{https://github.com/iSEE-Laboratory/RoGSplat}.
\end{abstract}

% *************************************
% ------------INTRODUCTION-------------
% *************************************
\section{Introduction}
\label{sec:intro}
Novel view synthesis is a widely discussed topic with various applications in video games, telepresence, sports broadcasting, and the metaverse. The past few years have witnessed remarkable progress in this domain, primarily due to the emergence of neural radiance fields (NeRF) \cite{mildenhall2020nerf} and 3D Gaussian Splatting (3DGS) \cite{3DGS}. However, human novel view synthesis still faces many challenges. First, a tedious per-subject optimization is typically required by existing methods to adapt to unseen human subjects, which significantly lowers their overall efficiency and practicability. Second, previous methods usually require dense input views or precise proxy geometry as input, thus their performance may degrade severely when only sparse views with limited overlappings are available. Third, the inherent intricate textures and self-occlusion in human characters introduce extra difficulty for high-fidelity free-view human rendering. 

To address these challenges, various NeRF-based methods are proposed to learn generalizable novel view synthesis for unseen human by leveraging pixel-aligned image features from multi-view images and human parametric template model (\eg, SMPL \cite{SMPL:2015}) as structural prior \cite{kwon2021neural, gao2023neural, kwon2023nip, hu2023sherf, Pan_2023_ICCV}. However, their performance is sensitive to the quality of the estimated template model, which is mostly unreliable for sparse multi-view images even ignoring the possible self-occlusion and depth ambiguities arising from varying human poses (see Figure~\ref{fig:teaser}). Moreover, these methods introduce extra computational overhead in addition to the costly MLP-based queries in NeRF, gaining generalizability at the expense of further degrading the rendering speed. While there are numerous works speeding up NeRF by adopting explicit representation such as voxels \cite{Liu2020NSVF,mueller2022instant,yu2022plenoxels}, tri-planes \cite{Chen2022ECCV,kplanes_2023}, and point clouds \cite{xu2022point,Hu_2023_CVPR,jiaxulearning}, developing generalizable NeRF that enables high-fidelity and efficient novel view synthesis remains an open problem.

3DGS-based methods have recently become mainstream for generalizable novel view synthesis because of the advantage of 3DGS in real-time rendering at high visual quality. Several pioneering works propose to achieve generalizability by incorporating depth estimation \cite{zheng2024gpsgaussian, charatan23pixelsplat, chen2024mvsplat,liu2025mvsgaussian}, but these methods are less effective for sparse multi-view images that can contain limitted overlappings or depth ambiguity derived from self-occlusions of human body (see Figure~\ref{fig:teaser}). Although a recent work \cite{pan2024humansplat} proposes to learn generalizable human 3DGS from a single-view image by utilizing a 2D multi-view diffusion model. However, it fails to offer high-fidelity novel view synthesis, especially when intricate garments and
accessories involved. 

In this paper, we present \model{}, a new generalizable human Gaussian Splatting approach that can robustly render high-fidelity novel views of unseen human subjects from sparse multi-view images. The core idea of our method lies in lifting SMPL vectices to more dense and accurate 3D points that accurately represent human body geometry for fine-grained 3D Gaussian regression. To enhance the robustness to the misalignment between SMPL model and multi-view images, we propose to leverage both the pixel-level features and voxel-level features for adaptive misalignment compensation, by which we are able to obtain 3D prior points well aligned with images. Based on the 3D prior points, we further introduce a coarse-to-fine pixel-wise Gaussian prediction strategy to regress fine-grained Gaussians that can effectively model high-frequency details.

In summary, the main contributions of our work are:

\begin{itemize}[leftmargin=2em] 
\setlength\itemsep{0.5em} 
    \item We present a novel generalizable human Gaussian Splatting method that enables robust high-fidelity novel view synthesis from sparse multi-view images. 

    \item We propose a method to effectively lift SMPL vertices to dense image-aligned 3D points, and introduce a coarse-to-fine pixel-wise Gaussian prediction strategy to enhance the effectiveness in modeling finer details. 
    
    \item Experiments on benchmark datasets demonstrate that our method outperforms state-of-the-art methods in novel view synthesis and cross-dataset generalization. 
\end{itemize}

% ********************************
% ------------RELATED-------------
% ********************************
\section{Related Work}
\label{sec:related}
\noindent \textbf{Neural implicit human reconstruction.} Early methods \cite{pifuSHNMKL19,zheng2020pamir,xiu2022icon} focus primarily on leveraging pixel-aligned image features to regress the signed distance field (SDF) under the supervision of ground-truth 3D models. Later, the emerging of neural radiance field (NeRF) \cite{mildenhall2020nerf} opens up a new paradigm to render high-fidelity human based only on sparse multi-view images \cite{peng2021neural,Zhao_2022_CVPR,shao2023tensor4d} or even a monocular image/video \cite{guo2023vid2avatar,jiang2022neuman,weng2022humannerf,hu2023sherf}. Although NeRF-based methods have demonstrated impressive results in animating \cite{liu2021neural,ARAH:2022:ECCV,li2022tava,li2023posevocab} and editing \cite{chen2022relighting,zhen2023relightable,Xu_2022_CVPR,Chen_2023_CVPR,NECA2024CVPR,chen2024meshavatar} human avatars, these methods typically fail to generalize to unseen subjects because of requiring per-subject optimization, and their rendering efficiency is low due to the inherent high computation to render each pixel. To achieve generalizable NeRF, a common solution is to construct feed-forward scene modeling conditioned on image-based features \cite{yu2021pixelnerf,wang2021ibrnet,lin2022enerf,chen2021mvsnerf}, but it is shown to be highly sensitive to large variation in pose and clothing. To address the issue, human priors such as SMPL \cite{SMPL:2015} and 3D skeleton keypoints are widely employed in recent works \cite{Pan_2023_ICCV,hu2023sherf,Zhao_2022_CVPR,kwon2021neural,Gao_2022_mpsnerf,mihajlovic2022keypointnerf}. On the other hand, there are various works for accelerating scene-specific NeRF \cite{mueller2022instant,Chen2022ECCV,yu2022plenoxels,kplanes_2023,TiNeuVox}. 
However, developing efficient yet generalizable NeRF remains a challenge.

\vspace{0.5em}
\qing{\noindent \textbf{Point-based rendering.} As a discrete and unstructured representation that can represent geometry with arbitrary topology, point element has been widely used in multi-view 3D reconstruction \cite{kopanas2022neural,lassner2021pulsar,wiles2020synsin,aliev2020neural,ruckert2022adop,xu2022point,lin2022learning,zheng2023pointavatar}. Recently, 3D Gaussian Splatting \cite{3DGS}, a novel scene representation allowing real-time and high-fidelity rendering by formulating point clouds as 3D Gaussians with learnable properties such as position, color, opacity and anisotropic covariance. Due to its superior performance in rendering quality and efficiency, a large number of works have been proposed to extend 3D Gaussian representation for widespread applications including dynamic scene modeling \cite{wu20234dgaussians,luiten2023dynamic,yang2023deformable3dgs,huang2023sc,yang2023gs4d}, human reconstruction \cite{li2024animatablegaussians,kocabas2024hugs,hu2024gaussianavatar,moreau2024human,hu2024gauhuman,jiang2024hifi4g,moon2024exavatar,xu2023gphm,chu2024gagavatar,wang2024v3,jiang2024dualgs}, and autonomous driving \cite{yan2024street,chen2024omnire,Zhou_2024_CVPR,zhou2024drivinggaussian}. However, these methods, although enable real-time inference, typically require either per-scene or per-frame optimization, preventing them from generalizing to unseen scenes. Recent attempts towards generalizable 3D Gaussian Splatting are basically built upon feed-forward regression of Gaussian parameters \cite{zheng2024gpsgaussian,kwon2024ghg,charatan23pixelsplat,szymanowicz24splatter,chen2024mvsplat,pan2024humansplat,prospero2024gstprecise3dhuman,jnchen24hgm,chen2024generalizable,liu2025mvsgaussian}. In particular, GPS-Gaussian \cite{zheng2024gpsgaussian} achieves generalizable novel view synthesis of unseen humans by jointly learning an iterative stereo-matching based depth estimation along with a Gaussian parameter regression, while GHG \cite{kwon2024ghg} regresses the 3D Gaussian parameters on the 2D UV space of a human template. However, these methods are usually less effective in handling sparse views with limited overlappings, from which depth can not be effectively estimated.} 

% ********************************
% ------------METHOD-------------
% ********************************
\input{figs/method}
\section{Method} 
Figure~\ref{fig:method} presents the overview of our method. As shown, given sparse views of a subject, our goal is to estimate prior points that well represent the human body geometry, and then regress 3D Gaussians in a feed-forward manner to gain generalizability and eliminate the need for per-subject optimization. Specifically, our method is comprised of two components, \ie, image-aligned human points prediction and coarse-to-fine pixel-wise Gaussian regression, with the former for prior points estimation and the latter for Gaussian regression. Below we describe our method in detail. 

\subsection{Image-aligned Human Points Prediction} 
\label{sec:initialization}
As shown in Figures~\ref{fig:teaser} and \ref{fig:ablate_smpl}, the SMPL model fitted from sparse views is typically unreliable due to limited overlappings. To get reliable yet dense prior points for Gaussian regression, we develop image-aligned human points prediction. Considering that SMPL model contains only sparse points, we employ the Snowflake Point Deconvolution (SPD) network \cite{xiang2023SPD,xiang2021snowflakenet} 
that is suitable for refining and densifying the estimated SMPL points $\mathbf{P}$ due to the ability to learn local geometric characteristics. Akin to \cite{xiang2023SPD,xiang2021snowflakenet}, we employ two SPD steps. To enhance its effectiveness and robustness, we integrate it with both pixel-level and voxel-level features to get the output human points by:
% , where the first aims to refine points with a upsampling factor of 1, while the second with a upsampling factor of 4 for point densification
\begin{equation}
\label{eq:up}
    \mathbf{P}^{o} = \mathcal{F}_{SPD}(\mathbf{P}, f_p, f_v),
\end{equation}
where $\mathbf{P}^{o}$ denotes the output points. $f_p$ and $f_v$ represent the utilized pixel-level and voxel-level features, respectively. 
% \xiao{We combine them by perform simple feature concatenation as they are both per-point related and share the same number of channels.}
 
\vspace{0.5em}
\noindent \textbf{Pixel-level features.}
To obtain projection-aware encoding that captures fine-grained local human details, while promoting the learning of spatial relationships between points and also the semantic information, we choose to adopt pixel-level features. To get the features, we first feed source multi-view images $\{ I_m | m=1,2,...,M\}$ into a U-Net based feature extractor $\mathcal{F}_{img}$ to obtain the corresponding feature maps $f_m = \mathcal{F}_{img}(I_m)$. Next, we conduct feature projection for these features, which consists of point projection and multi-view feature aggregation. Specifically, we first project the SMPL points $\mathbf{P}$ onto the multi-view feature maps to fetch points' feature embeddings $z_{m}= \Pi(f_m,\mathbf{P})$, where $\Pi$ denotes projection. We then employ a visibility-aware aggregation strategy \cite{wang24ICLR,Zhao_2022_CVPR,kwon2024ghg} to obtain pixel-level features $f_{p}$ by:
\begin{equation}
    f_{p} = \sum_{m=1}^M w_{m} \cdot z_{m},
\end{equation}
where $w_{m}$ denotes normalized visibility.

\vspace{0.5em}
\noindent \textbf{Voxel-level features.}
While pixel-level features provide projection-aware local details for human subject, they fail to represent accurate human geometry due to the lack of 3D geometric information. Therefore, we propose to adopt voxel-level features to complement the limitation of pixel-level features. We first feed the SMPL depth under the $m$-th view, denoted as $D_{m}$, and the related image $I_m$ into a U-Net based depth refiner network $\mathcal{F}_{d}$ to obtain the refined depth map $\hat{D}_{m}$ and corresponding depth features $H_{m}$:
\begin{equation} 
\label{eq:refine}
    (\hat{D}_{m}, H_{m}) = \mathcal{F}_{d}(D_{m}, I_m).
\end{equation}
Considering that directly unprojecting depth maps to 3D points is often too noisy to get reliable geometric information. We propose to voxelize them for producing a coarser but cleaner voxel feature volume that provides spatially aligned information. Specifically, we unproject the depth maps together with the corresponding features to 3D space, and then aggregate them to yield 3D points and per-point features. Next, we employ a sparse 3D convolution network \cite{Graham_2018_CVPR,Liu_2015_CVPR} to fuse nearby features. The final voxel-level features $f_{v}$ are obtained via tri-linear interpolation as follows:
\begin{equation} 
\label{voxel}
    f_{v} = \mathbf{T}(\mathbf{V},\mathbf{P}), 
\end{equation}
where $\mathbf{T}$ denotes tri-linear interpolation. $\mathbf{V}$ is the feature volume generated by the sparse convolution network. By integrating both pixel-level and voxel-level features, we can obtain image-aligned prior points that capture both fine-grained local 2D details and 3D-aware geometric information, which benefits generating high-fidelity human renderings, as demonstrated in Figure~\ref{fig:ablation}.

\vspace{0.5em}
\noindent \textbf{3D Gaussian regression.}
In the absence of available 3D information for supervision, we instead regress a set of coarse Gaussians. These can be rendered into images using differentiable rasterization, thereby providing effective supervision for SPD network. Each Gaussian point in 3D space is defined by attributes $\{ \mu, \mathbf{R}, \mathbf{S}, \alpha, \mathbf{c} \}$, representing position, rotation, scaling, opacity, and color, respectively. We utilize the prior points $\mathbf{P}^{o}$ as the positions of the 3D Gaussians, and then regress the remaining four properties based on both point features $\mathbf{q}$ from the last layer of SPD network and the pixel-level image features $f_{p}^o$ of prior points $\mathbf{P}^o$:
\begin{equation}
    \mathcal{G} = \mathcal{F}_{gs}(\mathbf{q},f_{p}^o),
\end{equation}
where $\mathcal{G}$ represents the estimated Gaussian properties. Note, although our utilized 3D prior points are significantly denser compared to the SMPL vertices, they are still insufficient to regress fine-grained Gaussians that allow for high-fidelity novel view synthesis, as demonstrated in Figure~\ref{fig:ablation}. However, the coarse Gaussians offer a rough yet image-aligned representation of geometric information, which is a basis of our subsequent coarse-to-fine pixel-wise Gaussian regression.

\input{figs/in_domain}
\input{tables/compare_with_nerf}

\subsection{Coarse-to-fine Pixel-wise Gaussian Regression} 
\label{sec:gaussian}
As described above, the predicted coarse Gaussians are insufficient for rendering high-fidelity novel views. To render high-fidelity images with fine-grained details, existing methods usually regress pixel-wise Gaussians based on depth information \cite{zheng2024gpsgaussian}. However, these methods basically fail to deal with sparse views that do not allow reliable depth estimation. Unlike previous methods, we find that our coarse Gaussians derived from prior points can help produce robust depth maps, which inspires us to develop an effective coarse-to-fine pixel-wise Gaussian regression.

With the coarse Gaussians derived from the prior points, we first rasterize them to generate per-view depth maps, which are then similarly refined via Eq. \eqref{eq:refine} to produce image-aligned depth maps. Then, we unproject these depth maps into 3D space to obtain pixel-wise points $\mathbf{P}^{'}$ that capture high-frequency details and more accurate human geometry. To further enhance the robustness of these points, we follow previous methods \cite{peng2021neural,shi2020pv,Peng2020ECCV} to utilize the prior points and the corresponding point features $\mathbf{q}$ to generate a latent feature volume $\mathbf{V}^{'}$ via sparse convolution network, and then interpolate the voxel-level features $f_{v}^{'}$ for pixel-wise points:
\begin{equation}
    f_{v}^{'} = \mathbf{T}(\mathbf{V}^{'},\mathbf{P}^{'}),
\end{equation}
where $\mathbf{T}$ denotes tri-linear interpolation and $\mathbf{V}^{'}$ represents the feature volume generated by sparse convolution network. This interpolated feature is then fed into a tiny MLP to predict offset $\delta$ of each point:
\begin{equation}
\delta = \mathcal{F}_{\delta}(f_{v}^{'}).
\end{equation}
The final points $\mathbf{P}^{g}$ are computed as $\mathbf{P}^{g} = \mathbf{P}^{'} + \delta$. Once the final points are obtained, we employ additional MLPs to predict pixel-wise 3D Gaussians:
\begin{equation}
    \mathcal{G}^{'} = \mathcal{F}_{gs}^{'}(\hat{f}_{p}^{g},f_{p}^g),
\end{equation}
where $\hat{f}_{p}^{g}$ and $f_{p}^g$ represent the pixel-level depth and image features of final points $\mathbf{P}^{g}$, respectively.

\input{figs/cross_domain}

\subsection{Training Details}
\label{sec:training}
To stabilize the training while obtaining better performance, we train our method in two stages. We first train the image-aligned human points prediction module to obtain stable coarse Gaussians. Then, we freeze this module and proceed to train the coarse-to-fine pixel-wise Gaussian regression.

\vspace{0.5em}
\noindent \textbf{Loss function for stage 1.}
We follow \cite{zheng2024gpsgaussian,3DGS} to use L1 loss and SSIM loss \cite{ssim} respectively indicated by $\mathcal{L}_{mae}$ and $\mathcal{L}_{SSIM}$, to measure the difference between the rendered output and the ground truth. Besides, we apply a mask loss \cite{hu2024gauhuman} to provide supervision for geometry learning:
\begin{equation}
    \mathcal{L}_{m} = || \mathbf{M}_{gt} - \mathbf{M}||_2,
\end{equation}
where $\mathbf{M}$ denotes the predicted target-view mask. $\mathbf{M}_{gt}$ indicates ground-truth mask. To constrain the depth refiner to produce meaningful estimations, we enforce its output $\hat{D}_m$ to be close to the input depth $D_m$ in valid areas:
\begin{equation}
    \mathcal{L}_{d} = || D_m[\mathcal{M}] - \hat{D}_m[\mathcal{M}] ||_1 ,
\end{equation}
where $\mathcal{M}$ means valid mask of depth map. Furthermore, inspired by \cite{depthanything}, we similarly define a semantic alignment loss to constrain the DINOv2 \cite{oquab2023dinov2} semantic features of the refined depth to be close to those of the input images:
\begin{equation}
    \mathcal{L}_{s} = || h_{img} - h_{depth} ||_1 ,
\end{equation}
where $h_{img}$ and $h_{depth}$ represent the semantic features of input images and refined depth.

The overall loss function for stage 1 is formulated as:
\begin{equation}
    \mathcal{L}_1 = \lambda_1\mathcal{L}_{mae} + \lambda_2\mathcal{L}_{SSIM} + \lambda_3 \mathcal{L}_{m} + \lambda_4 \mathcal{L}_{d}+\lambda_5 \mathcal{L}_{s},
\end{equation}
where we empirically set $\lambda_1 = 0.8$, $\lambda_2 = 0.2$,  $\lambda_3 = 1.0$, $\lambda_4 = 0.1$ and $\lambda_5 = 0.1$.

\vspace{0.5em}
\noindent \textbf{Loss function for stage 2.}
Since the coarse Gaussians are already approximately aligned with the images, we employ only the two photometric losses (\ie, $\mathcal{L}_{mae}$ and $\mathcal{L}_{SSIM}$) and the depth refinement loss $\mathcal{L}_{d}$ for training:
\begin{equation}
    \mathcal{L}_2 = \lambda_6\mathcal{L}_{mae} + \lambda_7\mathcal{L}_{SSIM}+\lambda_8\mathcal{L}_{d},
\end{equation}
where we set $\lambda_6 = 0.8$, $\lambda_7 = 0.2$ and $\lambda_8 = 0.1$.

\subsection{Implementation details}
\label{sec:training}
We first train the depth refiner for $20k$ iterations and then jointly train all modules for $200k$ iterations using a mini-batch size of 1 on an NVIDIA RTX 4090 GPU. The entire network is optimized using the AdamW optimizer with a learning rate of $1 \times 10^{-4}$. We set the voxel size for sparse convolution as $5mm \times 5mm \times 5mm$, and produce output feature volumes with downsampling factors of $2\times$, $4\times$, $8\times$, and $16\times$. Our method takes about 20 hours to train on the THuman2.0 dataset and 180ms for inference on $512 \times 512$ images.

% ********************************
% ------------EXPERIMENT----------
% ********************************
\section{Experiments}
\label{sec:experiments}

\input{tables/cross_data}
\subsection{Datasets and Metrics}
\label{sec:exp}
\qing{We evaluate our method on three benchmark human datasets including THuman2.0 \cite{tao2021function4d}, RenderPeople \cite{renderpeople}, and ZJU-MoCap~\cite{peng2021neural}. For THuman2.0, we select 400 subjects for training while the remaining 125 subjects for testing. As for RenderPeople, we utilize the version processed by \cite{hu2023sherf}, which consists of 482 subjects. We use 340 subjects for training and the others for testing. For ZJU-MoCap, we select 7 subjects for training and the other 3 subjects for testing. We also employ one real-world dataset collected by \cite{zheng2024gpsgaussian}. Note, we adopt \cite{easymocap} to estimate SMPL model parameters using fixed four-view (\ie, front, back, left, and right) images for all datasets. Similar to \cite{zheng2024gpsgaussian}, we employ PSNR, SSIM \cite{ssim}, and LPIPS \cite{Zhang_2018_CVPR} as metrics to quantitatively evaluate the rendering quality on the entire image.
}

\input{tables/compare_with_gs}

\input{figs/ablate_smpl}
\input{figs/ablation}

\subsection{Comparison with State-of-the-art Methods}
\noindent \textbf{{Baselines.}} We compare our method against four NeRF-based generalizable human rendering methods, including NHP \cite{kwon2021neural}, GP-NeRF  \cite{chen2022gpnerf}, TransHuman \cite{Pan_2023_ICCV} and SHERF \cite{hu2023sherf}, on THuman2.0~\cite{tao2021function4d}, RenderPeople~\cite{renderpeople}, and ZJU-MoCap~\cite{peng2021neural} datasets, and two generalizable human Gaussian Splatting methods including GPS-Gaussian \cite{zheng2024gpsgaussian} and GHG \cite{kwon2024ghg} only on THuman2.0 dataset due to following reasons: (i) GPS-Gaussian relies on RGB-D data rendered from 3D model for training while we only have available 3D models in THuman2.0. (ii) The training code of GHG \cite{kwon2024ghg} has not been released, we instead use their publicly-available pre-trained model trained on THuman2.0 dataset. For fair comparison, we retrain all baselines (except GHG) under the same training setting with 4 sparse input views, except GPS-Gaussian because it requires at least 6 views to train.

\input{tables/ablation}

\vspace{0.5em}
\noindent \textbf{In-domain generalization comparison.} Tables \ref{table:com_nerf} and \ref{table:com_gs} quantitatively compare our method with others on in-domain generalization. As shown, our approach clearly outperforms NeRF-based methods on THuman2.0, RenderPeople, ZJU-MoCap datasets, and exhibits superiority over other Gaussian Splatting based methods on THuman2.0 dataset. Figure~\ref{fig:in_domain} further qualitatively demonstrates that our method is able to generate high-fidelity human novel view renderings with well-preserved geometry and texture details. Note, although GHG also utilizes human template to serve as geometry prior like us, it does not consider the possible misalignment between template model and sparse-view images, and thus may induce results with inaccurate geometry. As for GPS-Gaussian, due to the difficulty to infer reliable depth from sparse views with limited overlapping, this method tends to produce incomplete human renderings. 
% Please see the supplementary material for more visual comparisons on novel view synthesis in terms of both images and videos.

\vspace{0.5em}
\noindent \textbf{Cross-domain generalization comparison.}
To evaluate the cross-domain generalizability of our method, we train our method and the compared methods on THuman2.0 dataset and then perform comparison on three challenging datasets including RenderPeople~\cite{renderpeople}, ZJU-MoCap~\cite{peng2021neural}, and a real-world dataset collected by \cite{zheng2024gpsgaussian}, without any test-time optimization. As shown in Table~\ref{table:cross_data}, our method achieves the best performance on all the three metrics, manifesting its advantage in cross-domain generalization. In addition, by comparing the visual results in Figure \ref{fig:cross_domain}, it is clear that our method is able to produce renderings with more accurate geometry and finer details.

\input{figs/pcd}

\subsection{More Analysis}
\noindent \textbf{Ablation studies.} Here we conduct ablation studies to validate the effectiveness of our pixel/voxel-level features, coarse-to-fine pixel-wise Gaussian regression, and offset estimator. Table~\ref{table:ablation} presents the numerical results of ablation studies on THuman2.0 dataset, where we can see that each of the four components has a clear contribution to the success of our method. Figure~\ref{fig:ablation} further presents visual demonstration. As shown, the adoption of the pixel- and voxel-level features benefits more accurate geometry estimation, while the coarse-to-fine pixel-wise Gaussian regression enables our method to capture finer details. The offset estimator helps obtain clearer rendering by avoiding incorrect Gaussian positioning. 
% Please see the supplementary material for analysis on different numbers of input views and other possible alternatives for Gaussian position prediction (\eg, predicting position or depth using Gaussian regressor) .

\vspace{0.5em}
\noindent \textbf{Robustness to inaccurate SMPL.} We also validate the robustness of our method to inaccurate SMPL models in Figure~\ref{fig:ablate_smpl}, where the top row gives the results of different methods produced with the SMPL model fitted from sparse views, while results on the bottom are produced from GT SMPL model provided in  dataset. As shown, despite that the fitted SMPL model on the top is clearly inaccurate, our method still produces high-fidelity result close to the one produced with GT SMPL. In contrast, the use of GT SMPL leads to significant performance increase in results of other compared methods. Figure~\ref{fig:pcd} further compares our estimated prior points and the GT SMPL points. As shown, our prior points are denser and capture accurate human geometry conformed with the GT SMPL points, effectively eliminating the misalignment between fitted SMPL points and the images.

\input{figs/failure_case}

\vspace{0.5em}
\noindent \textbf{Limitations.}
As shown in Figure~\ref{fig:failure_case}, our method may fail to handle loose clothing, from which it is very difficult to get a sufficient number of prior points. Besides, it may also struggle to recover high-frequency finer details for face and hands. Additionally, our method cannot animate or edit the reconstructed human models, and is designed to receive only fixed-view images as input. 

%\xiao{Finally, our method only support fixed-view images as input limiting overall practicality.} 

% ********************************
% ------------DISCUSSION----------
% ********************************
\section{Conclusion}
\label{conclusion}
We have presented \model{}, a robust generalizable human Gaussian Splatting that enables high-fidelity novel views from very sparse views. The key idea lies in lifting SMPL vertices to dense image-aligned 3D points representing human body geometry, and then performing point-based Gaussian regression. To this end, we present image-aligned human points prediction that integrates both pixel-level and voxel-level features to adaptively compensate the misalignment between the SMPL model and the multi-view images. With the obtained image-aligned points, we then introduce a coarse-to-fine pixel-wise Gaussian regression strategy to enhance the ability to model finer details. Experiments show that our method outperforms state-of-the-arts in photorealistic novel view synthesis and cross-dataset generalization.

\vspace{0.5em}
\noindent \textbf{Acknowledgement.} This work was supported by the National Natural Science Foundation of China (62471499), the Guangdong Basic and Applied Basic Research Foundation (2023A1515030002).

{\small
\bibliographystyle{ieeenat_fullname}
\bibliography{11_references}
}

\ifarxiv \clearpage \appendix \input{12_appendix} \fi

\end{document}

%% file: _constants.tex
\def\paperID{307}
\def\confName{CVPR}
\def\confYear{2025}

\newif\ifreview 
\newif\ifarxiv \newcommand{\arxiv}{\arxivtrue}
\newif\ifcamera 
\newif\ifrebuttal 
\newcommand{\model}{RoGSplat}

\newcommand{\eg}{e.g.}

%% file: cvpr_header.tex
\ifreview \usepackage[review]{cvpr} \fi
\ifarxiv \usepackage[pagenumbers]{cvpr} \fi
\ifrebuttal \usepackage[rebuttal]{cvpr} \fi
\ifcamera \usepackage{cvpr} \fi

\input{_macros}  % Add packages to _macros.tex

\usepackage{xr-hyper}

\makeatletter
\newcommand*{\addFileDependency}[1]{
  \typeout{(#1)}
  \@addtofilelist{#1}
  \IfFileExists{#1}{}{\typeout{No file #1.}}
}

\makeatother

\definecolor{cvprblue}{rgb}{0.21,0.49,0.74}
\usepackage[pagebackref,breaklinks,colorlinks,citecolor=cvprblue]{hyperref}
\usepackage[capitalize]{cleveref}
\crefname{section}{Sec.}{Secs.}
\crefname{table}{Table}{Tables}
\crefname{figure}{Fig.}{Figs.}

%% file: _macros.tex
\usepackage{graphicx}	
\usepackage{amsmath}	
\usepackage{amssymb}	
\usepackage{booktabs}
\usepackage{times}
\usepackage{microtype}
\usepackage{epsfig}
\usepackage{caption}
\usepackage{float}
\usepackage{placeins}
\usepackage{color, colortbl}
\usepackage{stfloats}
\usepackage{enumitem}
\usepackage{tabularx}
\usepackage{xstring}
\usepackage{multirow}
\usepackage{xspace}
\usepackage{url}
\usepackage{subcaption}
\usepackage{xcolor}
\usepackage[hang,flushmargin]{footmisc}
\ifcamera \usepackage[accsupp]{axessibility} \fi

% ...

\ifarxiv  \fi

\newcommand{\R}[1]{{%
    \textbf{%
        \ifstrequal{#1}{1}{\textcolor{red}{R#1}}{%
        \ifstrequal{#1}{2}{\textcolor{blue}{R#1}}{%
        \ifstrequal{#1}{3}{\textcolor{magenta}{R#1}}{%
        \ifstrequal{#1}{4}{\textcolor{teal}{R#1}}{%
                           \textcolor{cyan}{R#1}%
        }}}}%
    }%
}}

%% file: symbols_format.tex
\newcommand{\rev}[1]{{{#1}}} 

%%%%%%%%%%%%%%%%%%%%%%%%%%%%%%%%%%%%%%%%%%%%%%%%%%%%%%%%%%%%%%%%%%%
%%%%%%%%%%%%%%%%%%%%%%% SYMBOLS %%%%%%%%%%%%%%%%%%%%%%%%%%%%%%%
%%%%%%%%%%%%%%%%%%%%%%%%%%%%%%%%%%%%%%%%%%%%%%%%%%%%%%%%%%%%%%%%%%%
\newcommand{\ba}{\mathbf{a}}
\newcommand{\bb}{\mathbf{b}}
\newcommand{\bc}{\mathbf{c}}
\newcommand{\bd}{\mathbf{d}}
\newcommand{\be}{\mathbf{e}}
\newcommand{\bff}{\mathbf{f}}
\newcommand{\bg}{\mathbf{g}}
\newcommand{\bh}{\mathbf{h}}
\newcommand{\bi}{\mathbf{i}}
\newcommand{\bj}{\mathbf{j}}
\newcommand{\bk}{\mathbf{k}}
\newcommand{\bl}{\mathbf{l}}
\newcommand{\bn}{\mathbf{n}}
\newcommand{\bo}{\mathbf{o}}
\newcommand{\bp}{\mathbf{p}}
\newcommand{\bq}{\mathbf{q}}
\newcommand{\br}{\mathbf{r}}
\newcommand{\bs}{\mathbf{s}}
\newcommand{\bt}{\mathbf{t}}
\newcommand{\bu}{\mathbf{u}}
\newcommand{\bv}{\mathbf{v}}
\newcommand{\bw}{\mathbf{w}}
\newcommand{\bx}{\mathbf{x}}
\newcommand{\by}{\mathbf{y}}
\newcommand{\bz}{\mathbf{z}}
\newcommand{\bA}{\mathbf{A}}
\newcommand{\bB}{\mathbf{B}}
\newcommand{\bC}{\mathbf{C}}
\newcommand{\bD}{\mathbf{D}}
\newcommand{\bE}{\mathbf{E}}
\newcommand{\bF}{\mathbf{F}}
\newcommand{\bG}{\mathbf{G}}
\newcommand{\bH}{\mathbf{H}}
\newcommand{\bI}{\mathbf{I}}
\newcommand{\bJ}{\mathbf{J}}
\newcommand{\bK}{\mathbf{K}}
\newcommand{\bL}{\mathbf{L}}
\newcommand{\bM}{\mathbf{M}}
\newcommand{\bN}{\mathbf{N}}
\newcommand{\bO}{\mathbf{O}}
\newcommand{\bP}{\mathbf{P}}
\newcommand{\bQ}{\mathbf{Q}}
\newcommand{\bR}{\mathbf{R}}
\newcommand{\bS}{\mathbf{S}}
\newcommand{\bT}{\mathbf{T}}
\newcommand{\bU}{\mathbf{U}}
\newcommand{\bV}{\mathbf{V}}
\newcommand{\bW}{\mathbf{W}}
\newcommand{\bX}{\mathbf{X}}
\newcommand{\bY}{\mathbf{Y}}
\newcommand{\bZ}{\mathbf{Z}}
\newcommand{\balpha}{\mbox{\boldmath$\alpha$}}
\newcommand{\bgamma}{\mbox{\boldmath$\gamma$}}
\newcommand{\bGamma}{\mbox{\boldmath$\Gamma$}}
\newcommand{\bmu}{\mbox{\boldmath$\mu$}}
\newcommand{\bphi}{\mbox{\boldmath$\phi$}}
\newcommand{\bPhi}{\mbox{\boldmath$\Phi$}}
\newcommand{\bSigma}{\mbox{\boldmath$\Sigma$}}
\newcommand{\bsigma}{\mbox{\boldmath$\sigma$}}
\newcommand{\btheta}{\mbox{\boldmath$\theta$}}

\newcommand{\mE}{\mathcal{E}}
\newcommand{\mF}{\mathcal{F}}
\newcommand{\mB}{\mathcal{B}}
\newcommand{\mV}{\mathcal{V}}
\newcommand{\mG}{\mathcal{G}}
\newcommand{\mM}{\mathcal{M}}
\newcommand{\mH}{\mathcal{H}}
\newcommand{\mL}{\mathcal{L}}
\newcommand{\mU}{\mathcal{U}}
\newcommand{\mC}{\mathcal{C}}
\newcommand{\mS}{\mathcal{S}}
\newcommand{\mR}{\mathcal{R}}
\newcommand{\mD}{\mathcal{D}}
\newcommand{\mO}{\mathcal{O}}
\newcommand{\mP}{\mathcal{P}}
\newcommand{\mT}{\mathcal{T}}
\newcommand{\mSl}{\mathcal{S}_l}
\newcommand{\mN}{\mathcal{N}}
\newcommand{\mDll}{\mathcal{D}_{l,l'}}

\newcommand{\ra}{\rightarrow}
\newcommand{\la}{\leftarrow}

\def\A{{\cal A}}
\def\B{{\cal B}}
\def\C{{\cal C}}
\def\D{{\cal D}}
\def\E{{\cal E}}
\def\F{{\cal F}}
\def\G{{\cal G}}
\def\H{{\cal H}}
\def\I{{\cal I}}
\def\J{{\cal J}}
\def\K{{\cal K}}
\def\L{{\cal L}}
\def\M{{\cal M}}
\def\N{{\cal N}}
\def\O{{\cal O}}
\def\P{{\cal P}}
\def\Q{{\cal Q}}
\def\R{{\cal R}}
\def\S{{\cal S}}
\def\T{{\cal T}}
\def\U{{\cal U}}
\def\V{{\cal V}}
\def\W{{\cal W}}
\def\X{{\cal X}}
\def\Y{{\cal Y}}
\def\Z{{\cal Z}}
\def\Re{{\mathbb R}}
\def\Cx{{\mathbb C}}
\def\Ze{{\mathbb Z}}
\def\Na{{\mathbb N}}
\def\ud{\mathrm{d}}
\def\eps{\varepsilon}
\def\dist{\textrm{dist}}

% define colors for editing. Remove before upload to arxiv
\definecolor{ZhanColor}{rgb}{0,0.6,0}
\newcommand{\zhan}[1]{{\color{ZhanColor} \textbf{#1}}}

\definecolor{Qingcolor}{rgb}{0,0,0}
\newcommand{\qing}[1]{{\color{Qingcolor} {#1}}}

\definecolor{Xiaocolor}{rgb}{0.6,0.6,0}
\newcommand{\xiao}[1]{{\color{Xiaocolor} \textbf{#1}}}

%% file: figs/method.tex
\begin{figure*}[t]
    \centering
    \includegraphics[width=\linewidth]{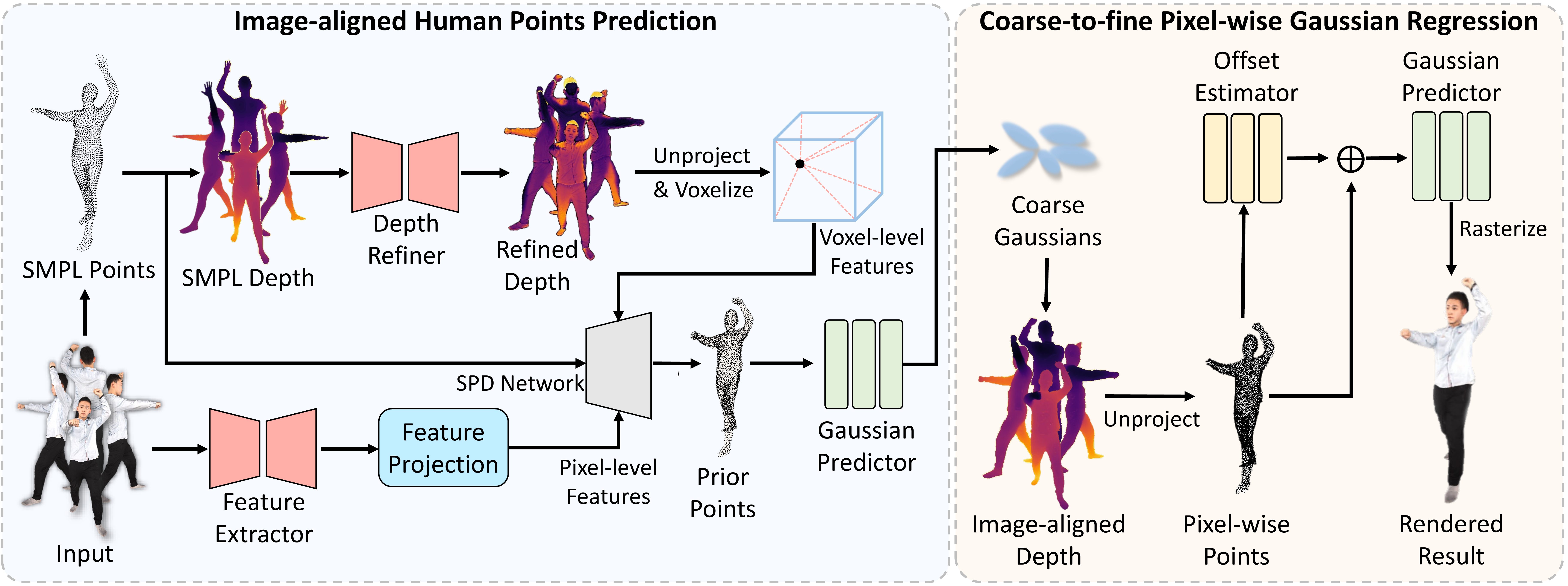}
    \vspace{-6mm}
    \caption{\qing{\textbf{Overview of \model{}. } We first fit the SMPL model from input sparse views, and then feed the SMPL depth into a depth refiner to get refined depth, from which we obtain voxel-level features. These features are then aggregated with pixel-level features extracted from source images, followed by the SPD network~\cite{xiang2023SPD, xiang2021snowflakenet} to generate dense image-aligned prior points for coarse Gaussian rasterization. To help model finer details, the image-aligned depth maps from coarse Gaussians are unprojected to yield finer pixel-wise points. These points are then refined by an offset estimator, and finally employed to regress fine-grained Gaussians. 
    }}
\label{fig:method}
\end{figure*}

%Given sparse multi-view images and an estimated SMPL model, we first learn a dual-scale human embedding that combines pixel and voxel level features. These features along with SMPL points, are fed into the SPD network~\cite{xiang2023SPD, xiang2021snowflakenet} to generate prior points for predicting coarse Gaussians. The image-aligned depth is then derived from coarse Gaussian rasterization. Depth unprojection yields pixel-wise points, which are refined through predicted per-point offset. Finally, we use these points to regress fine-grained Gaussians, enabling novel-view rasterization.

%% file: figs/in_domain.tex
\begin{figure*}[t]
    \centering
    \includegraphics[width=\linewidth]{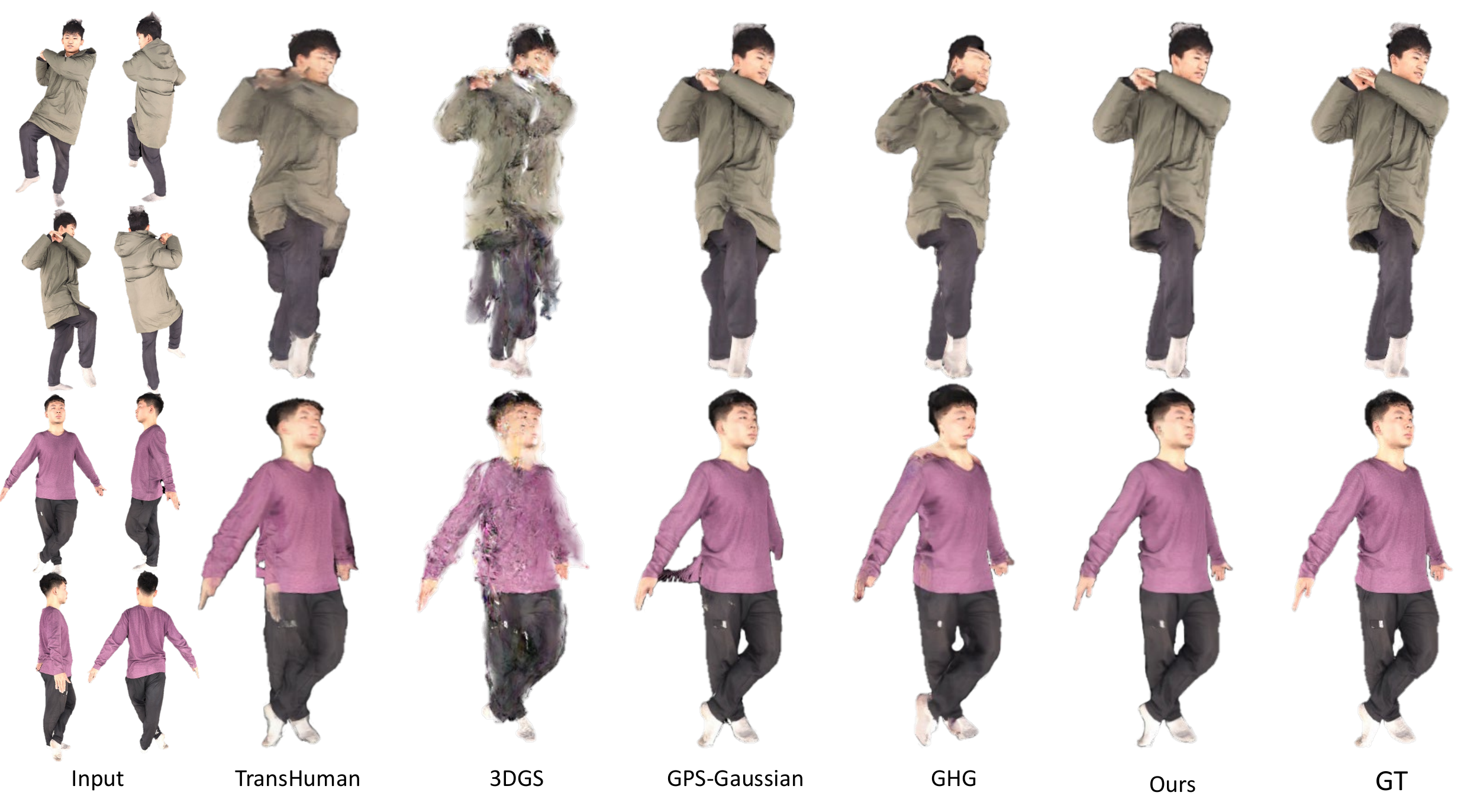}
    \vspace{-7mm}
    \caption{\textbf{Qualitative comparison of in-domain generalization on THuman2.0~\cite{tao2021function4d} dataset.}}
    \label{fig:in_domain}
\end{figure*}

%% file: tables/compare_with_nerf.tex
% Qualitative comparison
\begin{table*}[]
\centering
\caption{\textbf{Quantitative comparison of in-domain generalization with NeRF-based methods on THuman2.0~\cite{tao2021function4d}, RenderPeople~\cite{renderpeople}, and ZJU-MoCap~\cite{peng2021neural}.} Note, except for SHERF~\cite{hu2023sherf} which is evaluated on single view input, all other methods are evaluated on 4-view inputs.  }
\vspace{-1mm}
\resizebox{\linewidth}{!}
{
\begin{tabular}{l c c  c  c c c c c c c }
\toprule[1pt]
\multirow{2}{*}{Method}& \multicolumn{3}{c}{THuman2.0} & \multicolumn{3}{c}{RenderPeople} & \multicolumn{3}{c}{ZJU-MoCap} &\multirow{2}{*}{Param.}\\ 
 & PSNR $\uparrow$ & SSIM $\uparrow$ & LPIPS $\downarrow$ & PSNR $\uparrow$ & SSIM $\uparrow$ & LPIPS $\downarrow$ & PSNR $\uparrow$ & SSIM $\uparrow$ & LPIPS $\downarrow$&\\ 
\midrule
SHERF~\cite{hu2023sherf} &19.25 &0.8942 &0.1121 &23.38& 0.9379& 0.0767&27.81&0.9582&0.0526&56.4M \\
NHP~\cite{kwon2021neural} &25.74 &0.9356 &0.0748 &26.01& 0.9384& 0.0726&30.96&0.9644&0.0457&16.2M \\
GP-NeRF~\cite{chen2022gpnerf} &23.28 &0.9325 &0.0798 & 25.33&0.9326 &0.0792 &29.64&0.9563&0.0533&9.5M \\
TransHuman~\cite{Pan_2023_ICCV} &27.36 &0.9487 &0.0505 &26.37 &0.9451 & 0.0579&31.41&\textbf{0.9656}&0.0370&21.9M \\
Ours &\textbf{28.94} &\textbf{0.9615} &\textbf{0.0433} &\textbf{27.00} &\textbf{0.9530} &\textbf{0.0519}&\textbf{31.89}&0.9623&\textbf{0.0353}&12.2M \\
\bottomrule[1pt]
\end{tabular}
}
\vspace{-2mm}

\label{table:com_nerf}
\end{table*}

%We also report model size on the right column of table.

%% file: figs/cross_domain.tex
\begin{figure*}[t]
    \centering
    \includegraphics[width=\linewidth]{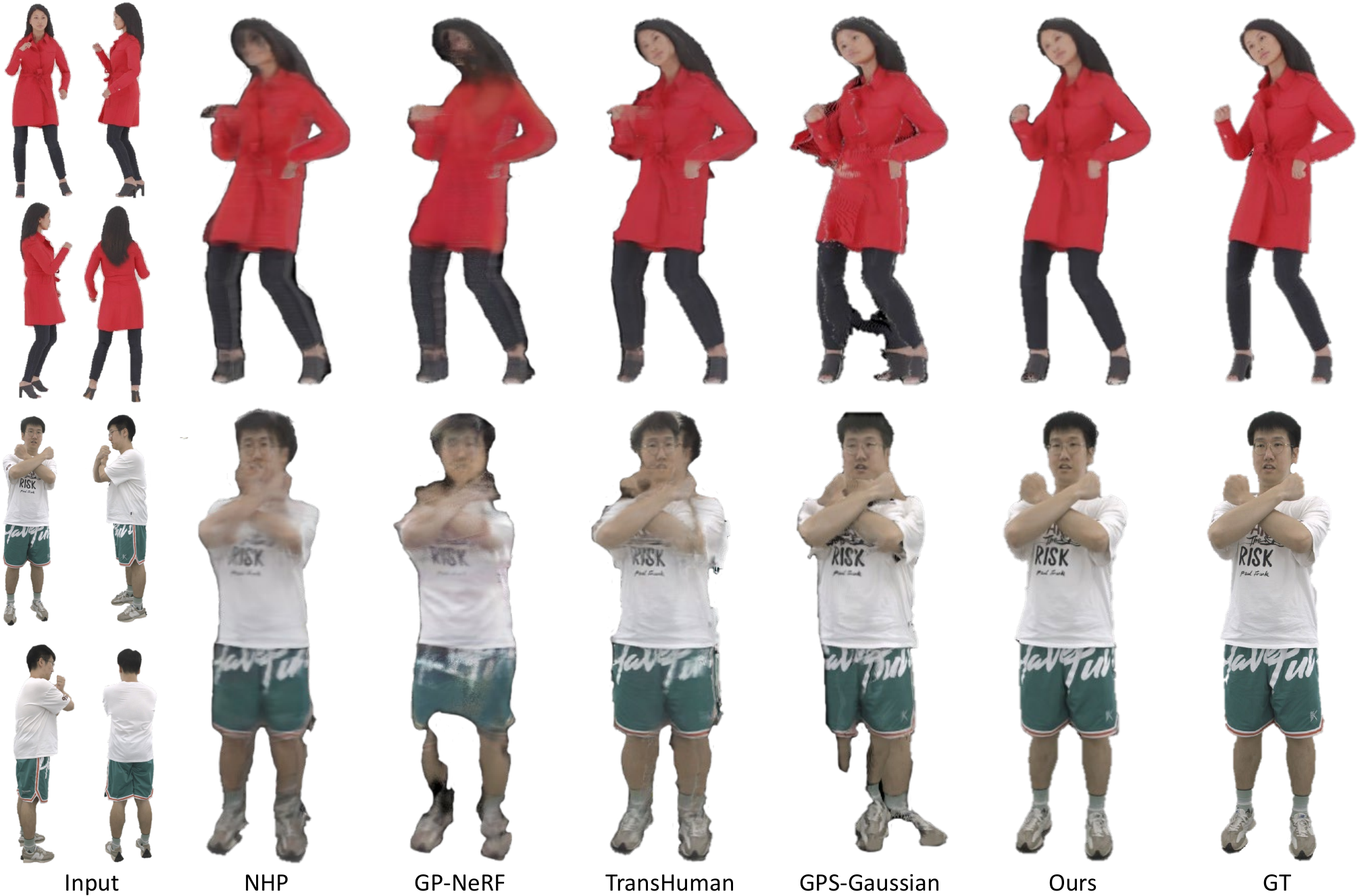}
    \vspace{-7mm}
    \caption{\textbf{Qualitative comparison of cross-domain generalization.} The results of each method here are produced by their models trained on the THuman2.0 dataset ~\cite{tao2021function4d}.}
    \label{fig:cross_domain}
\end{figure*}

    %We show the results trained with THuman2.0~\cite{tao2021function4d} and evaluated on ZJU-MoCap~\cite{peng2021neural} and real-world data~\cite{zheng2024gpsgaussian}.

%% file: tables/cross_data.tex
\begin{table*}[]
\centering
\caption{\textbf{Quantitative comparison of cross-domain generalization on RenderPeople~\cite{renderpeople}, ZJU-MoCap~\cite{peng2021neural}, and the real-world dataset collected by\cite{zheng2024gpsgaussian}.} Note, the results of each method here are produced by their models trained on the THuman2.0 dataset~\cite{tao2021function4d}.}
\vspace{-1mm}
\resizebox{\linewidth}{!}
{
\begin{tabular}{l c c c c c c c c c c}
\toprule[1pt]
\multirow{2}{*}{Method}&\multirow{2}{*}{View}&\multicolumn{3}{c}{RenderPeople} &\multicolumn{3}{c}{ZJU-MoCap} & \multicolumn{3}{c}{Real-world data}\\ 
 && PSNR $\uparrow$ & SSIM $\uparrow$ & LPIPS $\downarrow$& PSNR $\uparrow$ & SSIM $\uparrow$ & LPIPS $\downarrow$& PSNR $\uparrow$ & SSIM $\uparrow$ & LPIPS $\downarrow$\\ 
\midrule
NHP~\cite{kwon2021neural}&4&23.83&0.9276&0.0765&29.46 &0.9467 &0.0614&23.51&0.9182&0.0947\\
GP-NeRF~\cite{chen2022gpnerf}&4&23.13&0.9228&0.0852&26.78 &0.9398 &0.0693&20.09&0.8986&0.1130\\
TransHuman~\cite{Pan_2023_ICCV}&4&24.85&0.9347& 0.0640&29.57 &0.9473 &0.0583&24.60&0.9257&0.0744\\
GPS-Gaussian~\cite{zheng2024gpsgaussian}&6&25.11&0.9325&0.0682&29.06&0.9527&0.0464&21.55&0.9231&0.0847\\
Ours &4&\textbf{25.12}&\textbf{0.9380}&\textbf{0.0661}&\textbf{30.80}&\textbf{0.9596}&\textbf{0.0435}&\textbf{25.99}&\textbf{0.9452}&\textbf{0.0572}\\
\bottomrule[1pt]
\end{tabular}
}

\label{table:cross_data}
\end{table*}

%and evaluated on RenderPeople~\cite{renderpeople}, ZJU-MoCap~\cite{peng2021neural} and one real-world data~\cite{zheng2024gpsgaussian}

%% file: tables/compare_with_gs.tex
\begin{table}[]
\centering
\caption{\textbf{Quantitative comparison of in-domain generalization with Gaussian Splatting based methods on THuman2.0~\cite{tao2021function4d}.}}
\vspace{-1mm}
\resizebox{\linewidth}{!}
{

\begin{tabular}{l c c c c}
\toprule[1pt]
\multirow{1}{*}{Method}& \multicolumn{1}{c}{PSNR$\uparrow$}& \multicolumn{1}{c}{SSIM $\uparrow$}& \multicolumn{1}{c}{ LPIPS $\downarrow$}&\multicolumn{1}{c}{Param.}\\ 
\midrule
3DGS~\cite{3DGS}&22.07 &0.9068 &0.0846&/ \\
GPS-Gaussian~\cite{zheng2024gpsgaussian} &27.79 &0.9603 &\textbf{0.0397}&5.1M\\
GHG~\cite{kwon2024ghg} &24.98 &0.9418 &0.0522&17.8M\\
Ours &\textbf{28.94} &\textbf{0.9615} &0.0433&12.2M\\
\bottomrule[1pt]
\end{tabular}
}

\label{table:com_gs}
\end{table}

%% file: figs/ablate_smpl.tex
\begin{figure}[t]
    \centering
    \includegraphics[width=\linewidth]{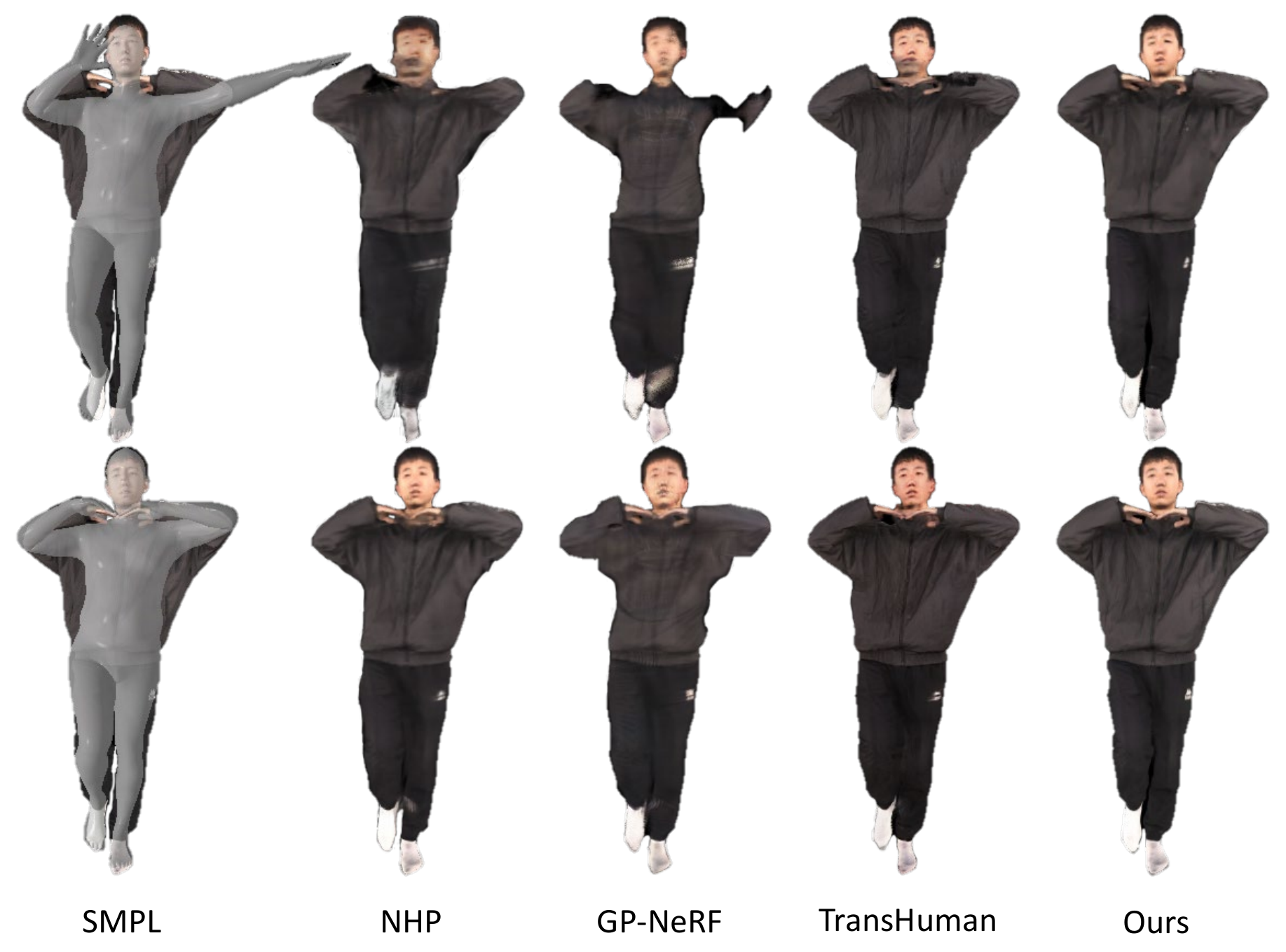}
    \vspace{-7mm}
    \caption{\textbf{Robustness to inaccurate SMPL.} Our method, unlike previous methods, can produce visually similar high-fidelity renderings from either inaccurate fitted SMPL or the GT SMPL.}
\label{fig:ablate_smpl}
\end{figure}

%We show the results using estimated SMPL from 4 views (top row) and ground truth SMPL (bottom row) provided in dataset to validate the robustness of our method.

%% file: figs/ablation.tex
\begin{figure*}[ht]  
  \centering    
  \captionsetup[subfigure]{labelformat=empty,labelsep=space}
  \begin{subfigure}[c]{0.135\linewidth}
        \centering
    	\begin{tikzpicture}[
		spy using outlines={color=red, rectangle, magnification=2,%connect spies,
			every spy on node/.append style={ size=0cm},
			every spy in node/.append style={ thick}}
		]
		\node{\includegraphics[width=1\textwidth, trim=110 30 110 30, clip]{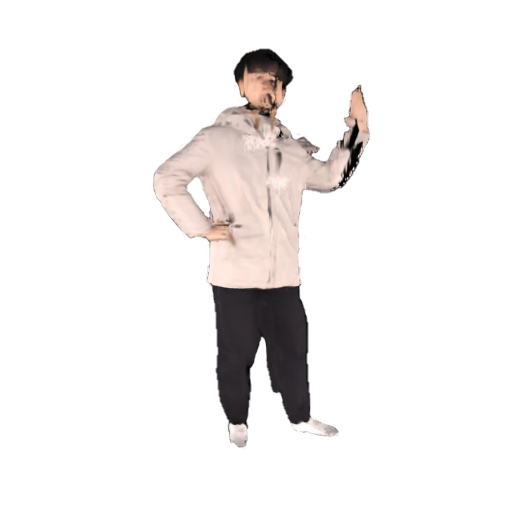}};
		\spy[width=1.8cm, height=1.4cm] on (-0.0, 1.8) in node at (-0.15, -2.8);
		\end{tikzpicture}
        \caption{\centering{w/o pix. \& vox. feat.}}
    \end{subfigure}
    \begin{subfigure}[c]{0.135\linewidth}
        \centering
    	\begin{tikzpicture}[
		spy using outlines={color=red, rectangle, magnification=2,%connect spies,
			every spy on node/.append style={ size=0cm},
			every spy in node/.append style={ thick}}
		]
		\node{\includegraphics[width=1\textwidth, trim=110 30 110 30, clip]{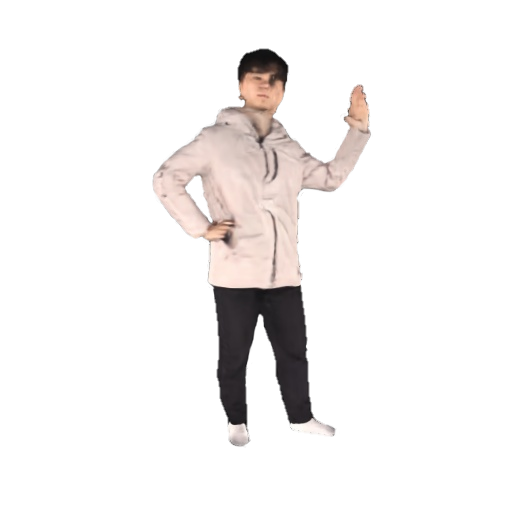}};
		\spy[width=1.8cm, height=1.4cm] on (-0.0, 1.8) in node at (-0.15, -2.8);
		\end{tikzpicture}
        \caption{\centering{w/o pixel. feat.}}
    \end{subfigure}
    \begin{subfigure}[c]{0.135\linewidth}
        \centering
    	\begin{tikzpicture}[
		spy using outlines={color=red, rectangle, magnification=2,%connect spies,
			every spy on node/.append style={ size=0cm},
			every spy in node/.append style={ thick}}
		]
		\node{\includegraphics[width=1\textwidth, trim=110 30 110 30, clip]{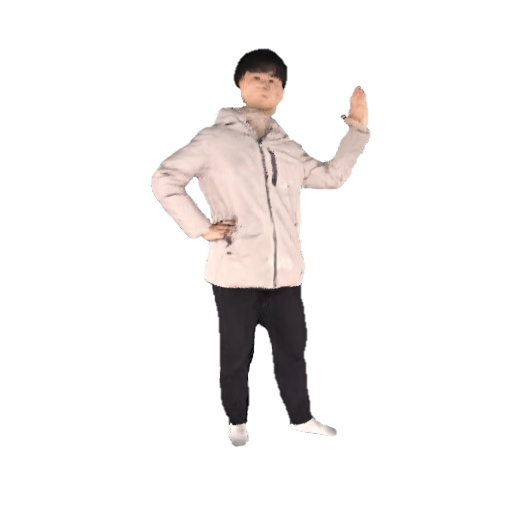}};
		\spy[width=1.8cm, height=1.4cm] on (-0.0, 1.8) in node at (-0.15, -2.8);
		\end{tikzpicture}
        \caption{\centering{w/o voxel. feat.}}
    \end{subfigure}
    \begin{subfigure}[c]{0.135\linewidth}
        \centering
    	\begin{tikzpicture}[
		spy using outlines={color=red, rectangle, magnification=2,%connect spies,
			every spy on node/.append style={ size=0cm},
			every spy in node/.append style={ thick}}
		]
		\node{\includegraphics[width=1\textwidth, trim=110 30 110 30, clip]{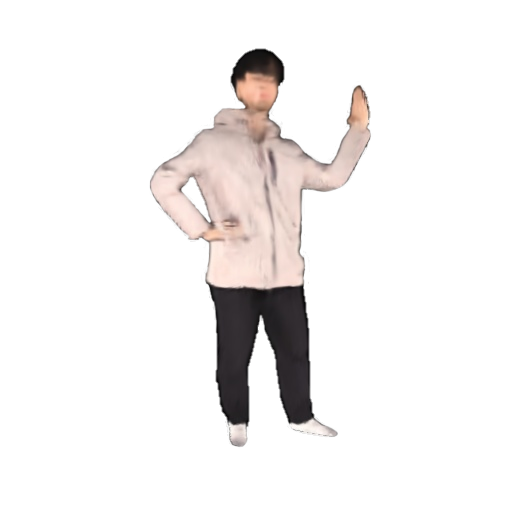}};
		\spy[width=1.8cm, height=1.4cm] on (-0.0, 1.8) in node at (-0.15, -2.8);
		\end{tikzpicture}
        \caption{\centering{w/o coarse-to-fine}}
    \end{subfigure}
    \begin{subfigure}[c]{0.135\linewidth}
        \centering
    	\begin{tikzpicture}[
		spy using outlines={color=red, rectangle, magnification=2,%connect spies,
			every spy on node/.append style={ size=0cm},
			every spy in node/.append style={ thick}}
		]
		\node{\includegraphics[width=1\textwidth, trim=110 30 110 30, clip]{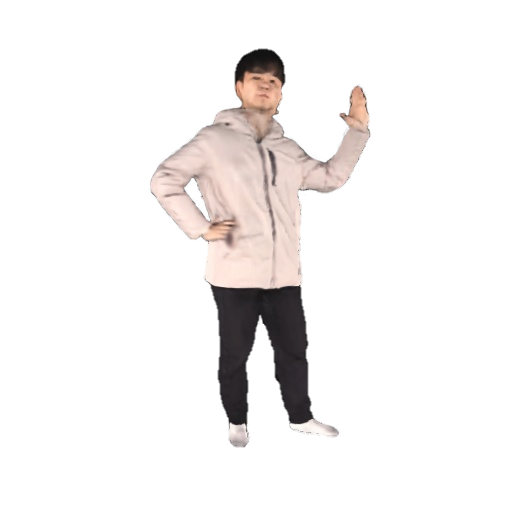}};
		\spy[width=1.8cm, height=1.4cm] on (-0.0, 1.8) in node at (-0.15, -2.8);
		\end{tikzpicture}
        \caption{\centering{w/o offset}}
    \end{subfigure}
    \begin{subfigure}[c]{0.135\linewidth}
        \centering
    	\begin{tikzpicture}[
		spy using outlines={color=red, rectangle, magnification=2,%connect spies,
			every spy on node/.append style={ size=0cm},
			every spy in node/.append style={ thick}}
		]
		\node{\includegraphics[width=1\textwidth, trim=110 30 110 30, clip]{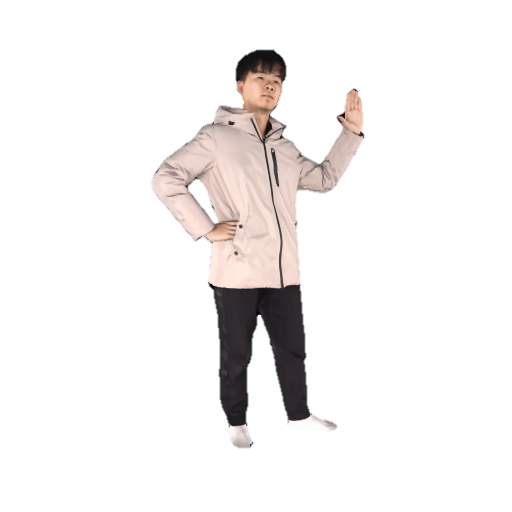}};
		\spy[width=1.8cm, height=1.4cm] on (-0.0, 1.8) in node at (-0.15, -2.8);
		\end{tikzpicture}
        \caption{\centering{full method}}
    \end{subfigure}
    \begin{subfigure}[c]{0.135\linewidth}
        \centering
    	\begin{tikzpicture}[
		spy using outlines={color=red, rectangle, magnification=2,%connect spies,
			every spy on node/.append style={ size=0cm},
			every spy in node/.append style={ thick}}
		]
		\node{\includegraphics[width=1\textwidth, trim=110 30 110 30, clip]{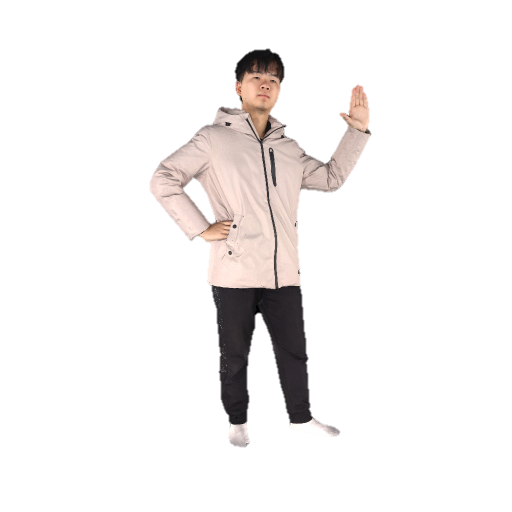}};
		\spy[width=1.8cm, height=1.4cm] on (-0.0, 1.8) in node at (-0.15, -2.8);
		\end{tikzpicture}
        \caption{\centering{ground truth}}
    \end{subfigure}
  
 \vspace{-3mm}
 \caption{\textbf{Ablation studies on THuman2.0 dataset.} ``w/o pix. \& vox. feat.'' refers to omitting both pixel-level and voxel-level features, while ``w/o pixel. feat.'' and ``w/o voxel. feat.'' correspond to ignoring pixel-level features and voxel-level features, respectively. ``w/o coarse-to-fine'' means we directly use coarse Gaussians for novel view synthesis, and ``w/o offset'' indicates removing the offset estimator.}
\label{fig:ablation}
\end{figure*}

%% file: tables/ablation.tex
\begin{table}[]
\centering
\caption{\textbf{Ablation studies on THuman2.0~\cite{tao2021function4d} dataset.} Note, ``baseline'' here indicates we directly feed originally fitted SMPL points into the SPD network for coarse Gaussian regression.}
\vspace{-1mm}
\resizebox{\linewidth}{!}
{
\begin{tabular}{l c c c}
\toprule[1pt]
\multirow{1}{*}{Method}& \multicolumn{1}{c}{PSNR$\uparrow$}& \multicolumn{1}{c}{SSIM $\uparrow$}& \multicolumn{1}{c}{ LPIPS $\downarrow$}\\ 
\midrule
baseline &23.28 &0.9301 &0.0975\\
w/o pixel \& voxel-level features &27.41 &0.9467 &0.0629\\
w/o pixel-level features &28.45 &0.9603 &0.0460\\
w/o voxel-level features &28.28 &0.9585 &0.0459\\
w/o coarse-to-fine Gaussian regress. &28.00 &0.9563 &0.0602\\
w/o offset estimator &28.56 &0.9611 &0.0439\\
full method&\textbf{28.94} &\textbf{0.9615} &\textbf{0.0433}\\
\bottomrule[1pt]
\end{tabular}
}

\label{table:ablation}
\end{table}

%``w/o pix. \& vox. feat.'' denotes that both pixel-level and voxel-level features are not employed, while ``w/o pixel. feat.'' and ``w/o voxel. feat.'' indicate omitting pixel-level feature and voxel-level feature, respectively. ``w/o coarse-to-fine'' means we directly use coarse Gaussians for novel view synthesis, and ``w/o offset'' indicates without offset estimator

%% file: figs/pcd.tex
\begin{figure}[!t]  
  \centering    
  \captionsetup[subfigure]{labelformat=empty,labelsep=space}
  \begin{subfigure}[c]{0.31\linewidth}		\includegraphics[width=1\textwidth, trim=80 40 80 55, clip]{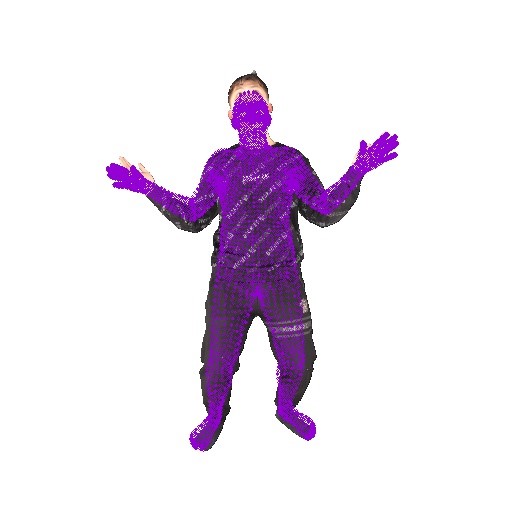}
    \caption{Fitted SMPL pcd}
\end{subfigure}
\begin{subfigure}[c]{0.31\linewidth}		         \includegraphics[width=1\textwidth, trim=80 40 80 55, clip]{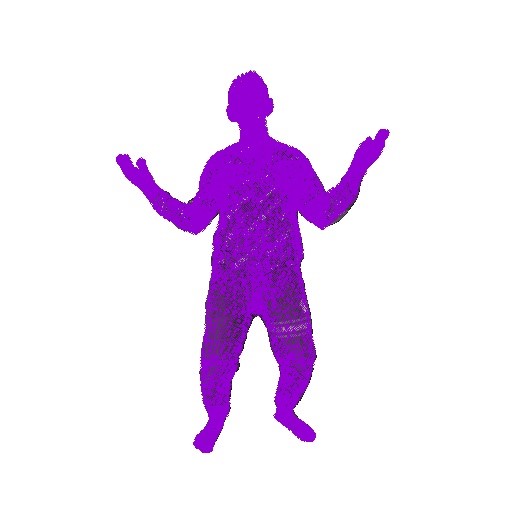}
    \caption{Our refined pcd}
\end{subfigure}
    \begin{subfigure}[c]{0.31\linewidth}		\includegraphics[width=1\textwidth, trim=80 40 80 55, clip]{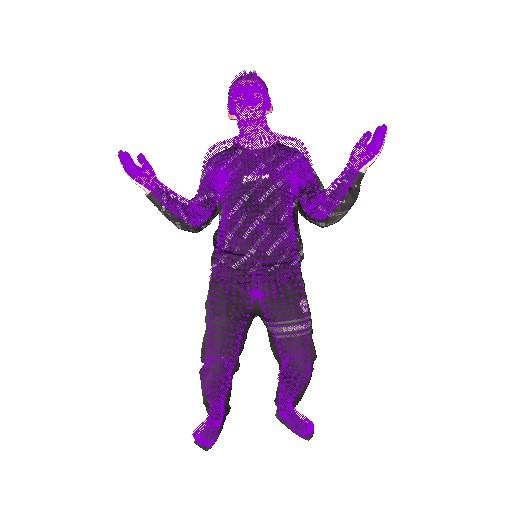}
    \caption{GT SMPL pcd}
    \end{subfigure}
 \vspace{-2mm}
 \caption{\textbf{Comparison of our estimated prior points and GT SMPL points.} Here we project estimated SMPL points, our predicted prior points as well as the GT SMPL points provided in dataset onto image for comparison.}
\label{fig:pcd}
\end{figure}

%% file: figs/failure_case.tex
\begin{figure}[!t]  
  \centering    
  \captionsetup[subfigure]{labelformat=empty,labelsep=space}
  \begin{subfigure}[c]{0.242\linewidth}		\includegraphics[width=1\textwidth, trim=130 80 120 60, clip]{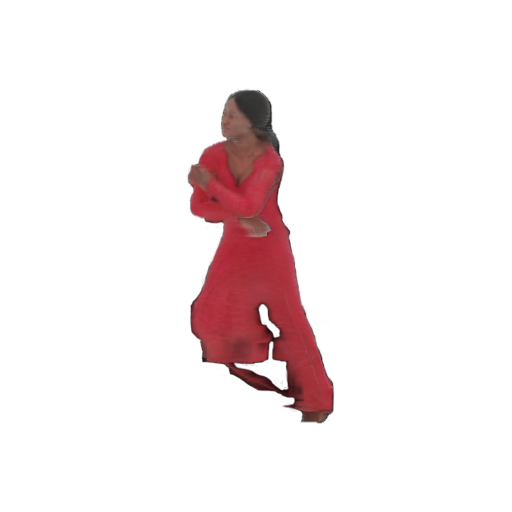}
    \caption{NHP}
\end{subfigure}
\begin{subfigure}[c]{0.242\linewidth}		         \includegraphics[width=1\textwidth, trim=130 80 120 60, clip]{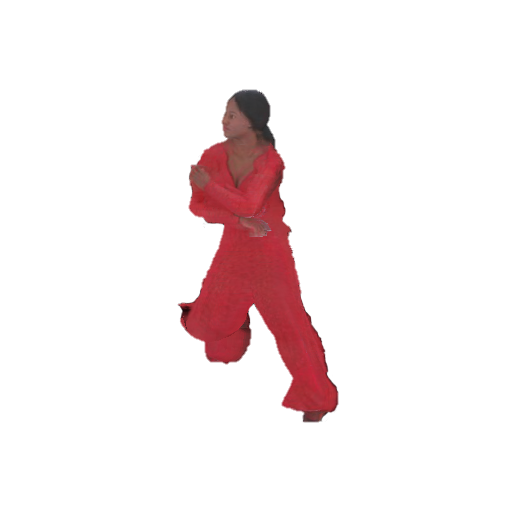}
    \caption{TransHuman}
\end{subfigure}
    \begin{subfigure}[c]{0.242\linewidth}		\includegraphics[width=1\textwidth, trim=130 80 120 60, clip]{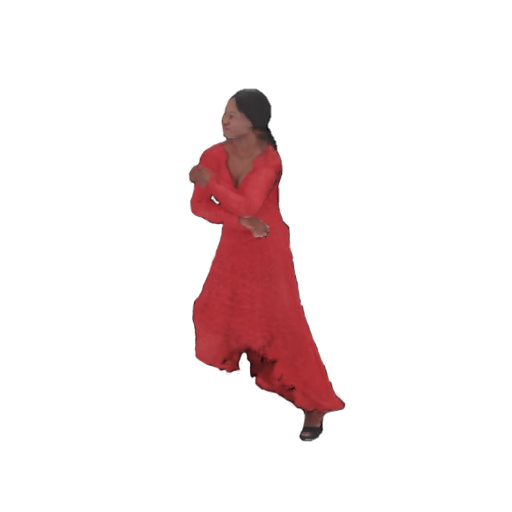}
    \caption{Ours}
    \end{subfigure}
    \begin{subfigure}[c]{0.242\linewidth}		\includegraphics[width=1\textwidth, trim=130 80 120 60, clip]{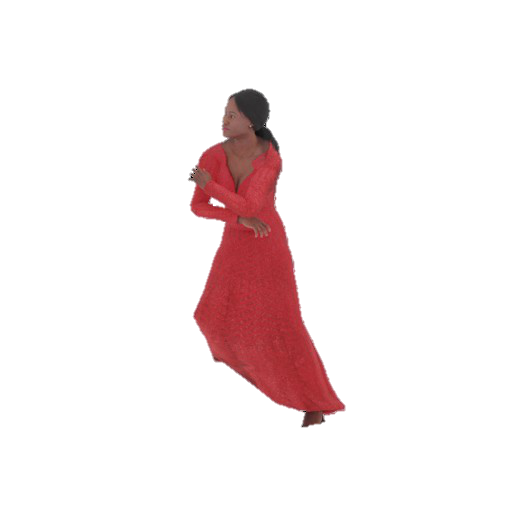}
    \caption{GT}
    \end{subfigure}
 \vspace{-2mm}
 \caption{\textbf{Failure case.} Our method fails to handle loose clothing.}
\label{fig:failure_case}
\end{figure}

%% file: 12_appendix.tex
\section{Implementation Details}
\noindent \textbf{Feature extractor and depth refiner.}
We employ the same network architecture for both the image feature extractor and the depth refiner, with differences only in their input and output channels. The feature extractor processes the input image and generates a 32-channel feature map while preserving the input image resolution. Conversely, the depth refiner takes both the original depth map and the image as input and produces a refined depth map as output. The network architecture is illustrated in Figure~\ref{fig:unet}. Each residual block in the network comprises two $3 \times 3$ convolutional layers, each followed by ReLU activation and group normalization.

\noindent \textbf{Sampling voxel-level features.}
Given a set of target points (estimated SMPL points $\mathbf{P}$ or pixel-wise points $\mathbf{P}^{'}$ in our work), we aim to sample their voxel-level features from a constructed feature volume to capture 3D-aware geometric information. To achieve this, we first construct a feature volume using sparse 3D points (unprojected points derived from refined depth or prior points $\mathbf{P}^o$ in our work). Inspired by \cite{peng2021neural,kwon2023nip}, we utilize SparseConvNet \cite{Graham_2018_CVPR, Liu_2015_CVPR} to diffuse the features of these sparse 3D points. The network architecture is detailed in Table~\ref{tab:sparse_conv}. Initially, we compute the 3D bounding box of the sparse points and divide it into small voxels, each measuring $5mm \times 5mm \times 5mm$, resulting in a volumetric representation. SparseConvNet processes this volumetric input using 3D sparse convolutions, diffusing the features of the sparse points into the surrounding 3D space and producing output features. The multi-scale outputs from the 5th, 9th, 13th, and 17th layers of SparseConvNet are resized and concatenated to form the final feature volume. Voxel-level features are then sampled from this output volume using tri-linear interpolation.

\noindent \textbf{Gaussian predictor.}
As shown in Figure~\ref{fig:gs}, our Gaussian predictor networks take point-related features as input, which include pixel-level image features and point features from the SPD network (or pixel-level depth features). These networks output Gaussian properties. Each predictor head is implemented as a 3-layer MLP, with each layer (except the final one) producing 256-dimensional features.

\noindent \textbf{Offset estimator.}
The architecture of the offset estimator is illustrated in Figure~\ref{fig:offset}. It takes the voxel-level feature $f_v^{'}$ of pixel-wise points as input and outputs per-point offset. Each layer generates 128-dimensional features with ReLU activations, except for the final layer, which employs Tanh activations.
\input{tables/supp/ablate_view}
\input{figs/ablate_view}
\input{figs/alternatives}
\input{tables/supp/alternatives}

\input{figs/unet}
\input{tables/supp/sp_conv}
\input{figs/gs}
\input{figs/offset}

\section{Additional Qualitative Results}
Figure \ref{fig:more_thu} showcases additional qualitative comparison of in-domain generalization results. Compared to all baseline methods, our method can preserve more reasonable geometry and high-fidelity appearance details. Figure  \ref{fig:more_cross} provides additional qualitative comparison of cross-domain generalization, where we show the results on RenderPeople~\cite{renderpeople,hu2023sherf} dataset using model trained on THuman2.0~\cite{tao2021function4d} dataset, demonstrating that our method outperforms others on cross-dataset generalization. For further results, please refer to our supplementary video.
\input{figs/thuman}
\input{figs/cross}
\section{Additional Ablation Studies}
\noindent \textbf{Ablation study on the number of input views.}
We evaluate our method with varying numbers of input views, as presented in Table~\ref{table:ablate_view} and Figure~\ref{fig:ablate_view}. The results indicate that performance improves as the number of input views increases, providing more observations.

\noindent \textbf{Ablation study on alternatives for Gaussian position prediction.}
To validate the effectiveness of our proposed coarse-to-fine pixel-wise Gaussian prediction method, which leverages refined prior 3D points to regress fine-grained 3D Gaussians, we compare it against two alternative approaches: jointly regressing Gaussian positions or depth maps alongside other Gaussian properties via Gaussian predictor, denoted as ``position'' and ``depth'' respectively. To ensure valid numerical results, we clamp the predicted position and depth values to the range $[0,1]$ and scale these values according to the bounding box of the fitted SMPL model. As shown in Table~\ref{table:alternatives} and Figure~\ref{fig:alternatives}, our proposed strategy demonstrates superior performance compared to the other two alternatives.

\input{tables/rebuttal/geo_ablation}
\noindent \textbf{Ablation studies on depth refiner and coarse Gaussian predictor.}
In table \ref{table:geo_ablation}, we conduct ablation studies on THuman2.0 dataset to quantitatively evaluate the contribution of our depth refiner and coarse Gaussian prediction, where we can see that the two geometry-related designs benefit our method to obtain better results.

\section{Influence of pose distribution in the training set.}
Voxel-level features depend on the pose distribution in the training set, which may adversely affect our generalizability to unseen poses. However, this influence is effectively mitigated by incorporating pixel-level features and performing coarse-to-fine Gaussian prediction, as demonstrated by the results on the RenderPeople dataset in Table \ref{table:rp}. As shown, although the RenderPeople dataset contains diverse challenging poses, our method still outperforms the compared methods, manifesting its advantage in generalizability.

\input{tables/rebuttal/rp}
\section{Ethics Statement}

The datasets utilized in our research are sourced from publicly available repositories, including THuman2.0~\cite{tao2021function4d}, RenderPeople~\cite{renderpeople,hu2023sherf}, ZJU-MoCap \cite{peng2021neural}, and one real-world data~\cite{zheng2024gpsgaussian}. Our research centers on the development of a method for free-viewpoint rendering of unseen human avatars, a technology poised to have significant implications in various domains, particularly within virtual environments such as the metaverse and video games. However, it is crucial to recognize the potential misuse and ethical concerns associated with this technology. The ability to manipulate digital representations of individuals, particularly in photorealistic and indistinguishable ways, raises legitimate concerns regarding privacy, identity theft, and the perpetration of fraudulent activities. In light of these considerations, we advocate for the responsible and ethical use of our research findings.

%% file: tables/supp/ablate_view.tex
\begin{table}[]
\centering

\resizebox{0.9\linewidth}{!}
{
\begin{tabular}{c c c c}
\toprule[1pt]
\multirow{1}{*}{Num of view}& \multicolumn{1}{c}{PSNR$\uparrow$}& \multicolumn{1}{c}{SSIM $\uparrow$}& \multicolumn{1}{c}{ LPIPS $\downarrow$}\\ 
\midrule
1 &21.31 &0.9123 &0.1026\\
2 &23.57 &0.9255 &0.0857\\
3 &26.32 &0.9478 &0.0530\\
4 &28.94 &0.9615 &0.0433\\
5 &30.98 &0.9711 &0.0341\\
\bottomrule[1pt]
\end{tabular}
}
\vspace{-2mm}
\caption{\textbf{Ablation study on the number of input views.} We train and test our method given different input views. The performance improves with more available observations.}
\label{table:ablate_view}
\end{table}

%% file: figs/ablate_view.tex
\begin{figure}[!t]  
  \centering    
  \captionsetup[subfigure]{labelformat=empty,labelsep=space}
  \begin{subfigure}[c]{0.19\linewidth}		\includegraphics[width=1\textwidth, trim=180 40 150 60, clip]{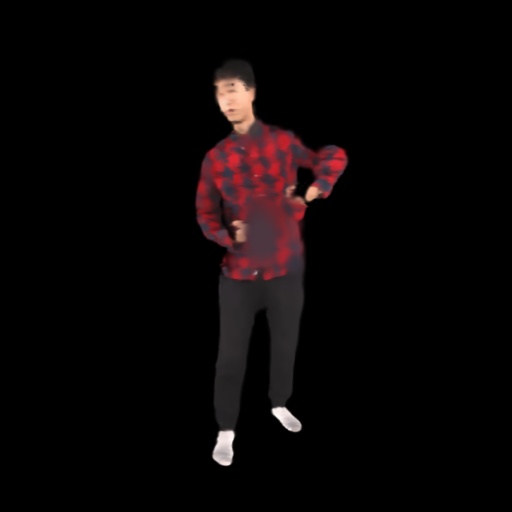}
    \caption{1 view}
\end{subfigure}
\begin{subfigure}[c]{0.19\linewidth}		\includegraphics[width=1\textwidth, trim=180 40 150 60, clip]{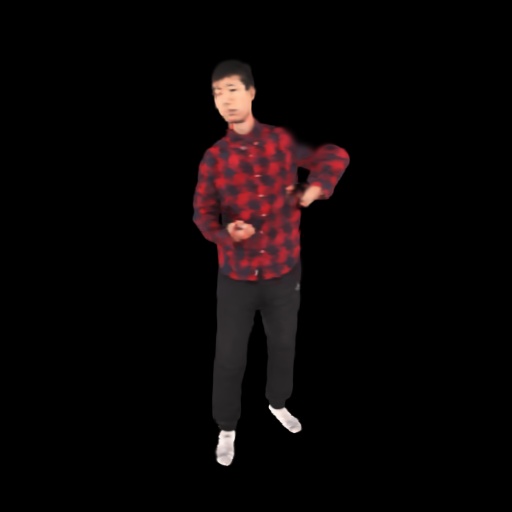}
    \caption{2 views}
\end{subfigure}
\begin{subfigure}[c]{0.19\linewidth}		\includegraphics[width=1\textwidth, trim=180 40 150 60, clip]{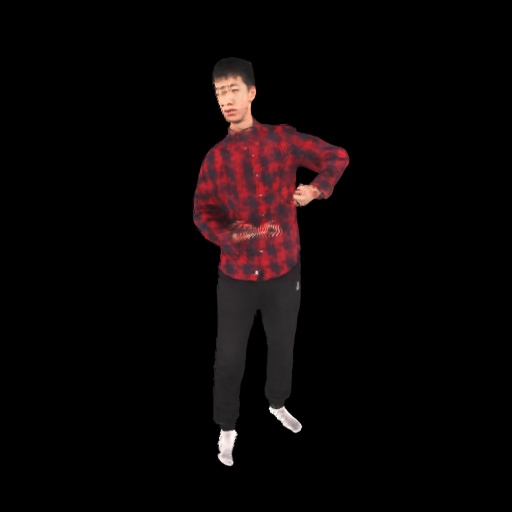}
    \caption{3 views}
\end{subfigure}
\begin{subfigure}[c]{0.19\linewidth}		\includegraphics[width=1\textwidth, trim=180 40 150 60, clip]{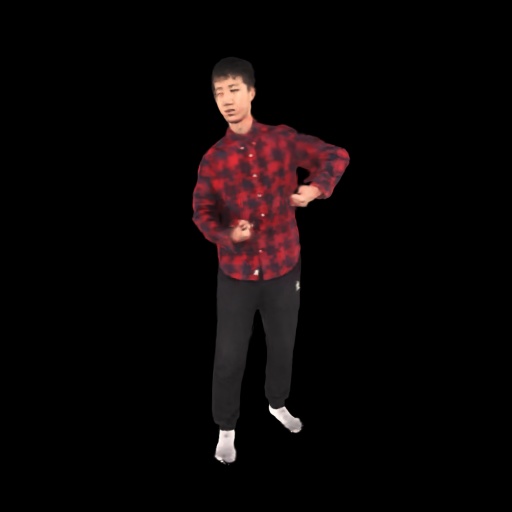}
    \caption{4 views}
\end{subfigure}
\begin{subfigure}[c]{0.19\linewidth}		\includegraphics[width=1\textwidth, trim=180 40 150 60, clip]{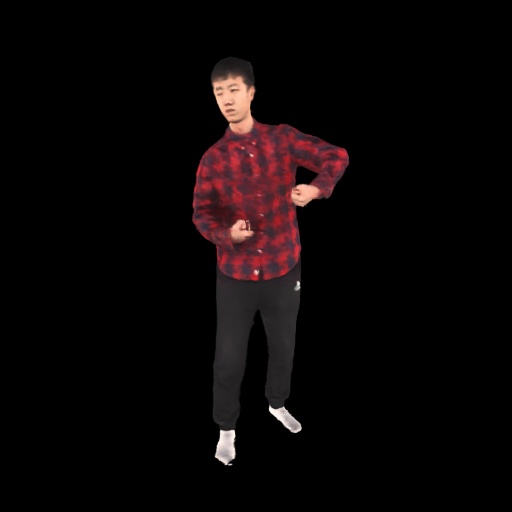}
    \caption{5 views}
\end{subfigure}
\vspace{-2mm}
 \caption{\textbf{Ablation study on the number of input views.}}
\label{fig:ablate_view}
\end{figure}

%% file: figs/alternatives.tex
\begin{figure}[!t]  
  \centering    
  \captionsetup[subfigure]{labelformat=empty,labelsep=space}
  \begin{subfigure}[c]{0.24\linewidth}		\includegraphics[width=1\textwidth, trim=160 40 150 60, clip]{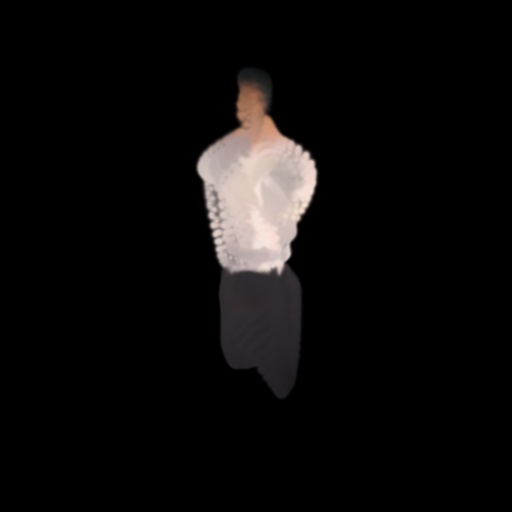}
    \caption{position}
  \end{subfigure}
  \begin{subfigure}[c]{0.24\linewidth}		\includegraphics[width=1\textwidth, trim=160 40 150 60, clip]{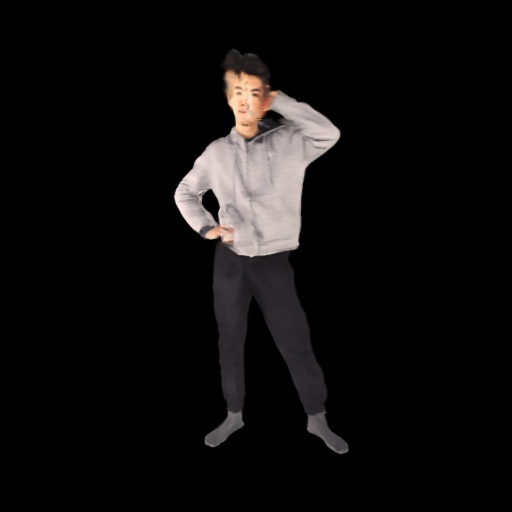}
    \caption{depth}
  \end{subfigure}
  \begin{subfigure}[c]{0.24\linewidth}		\includegraphics[width=1\textwidth, trim=160 40 150 60, clip]{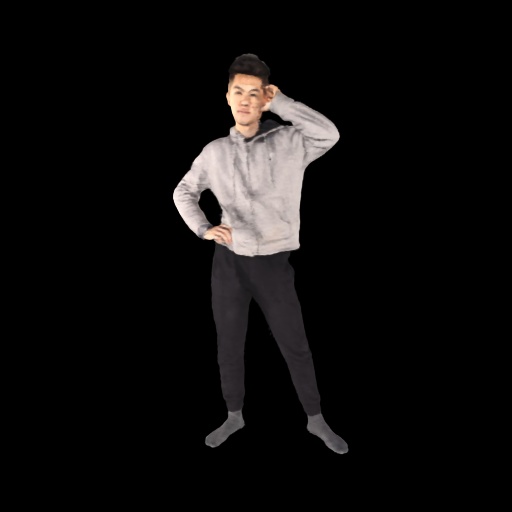}
    \caption{Ours}
  \end{subfigure}
  \begin{subfigure}[c]{0.24\linewidth}		\includegraphics[width=1\textwidth, trim=160 40 150 60, clip]{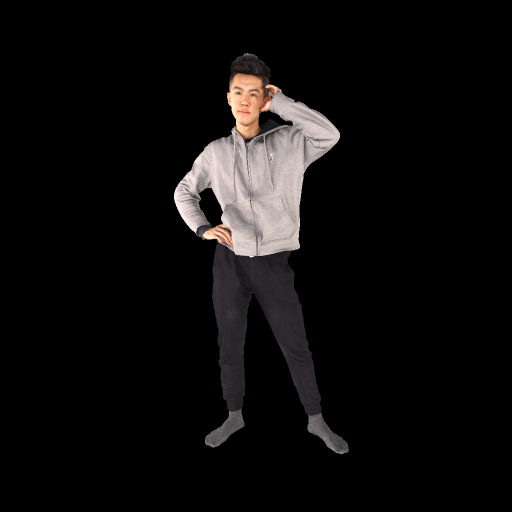}
    \caption{GT}
  \end{subfigure}
  \vspace{-2mm}
 \caption{\textbf{Ablation study on alternatives for Gaussian position prediction.}}
 
\label{fig:alternatives}
\end{figure}

%% file: tables/supp/alternatives.tex
\begin{table}[]
\centering

\resizebox{0.7\linewidth}{!}
{
\begin{tabular}{c c c c}
\toprule[1pt]
\multirow{1}{*}{Method}& \multicolumn{1}{c}{PSNR$\uparrow$}& \multicolumn{1}{c}{SSIM $\uparrow$}& \multicolumn{1}{c}{ LPIPS $\downarrow$}\\ 
\midrule
position &18.34 &0.8915 &0.1271\\
depth &24.25 &0.9405 &0.0666\\
Ours &28.94 &0.9615 &0.0433\\
\bottomrule[1pt]
\end{tabular}
}
\vspace{-2mm}
\caption{\textbf{Ablation study on alternatives for Gaussian position prediction.}}
\label{table:alternatives}
\end{table}

%% file: figs/unet.tex
\begin{figure*}[h]
    \centering
    \includegraphics[width=\linewidth]{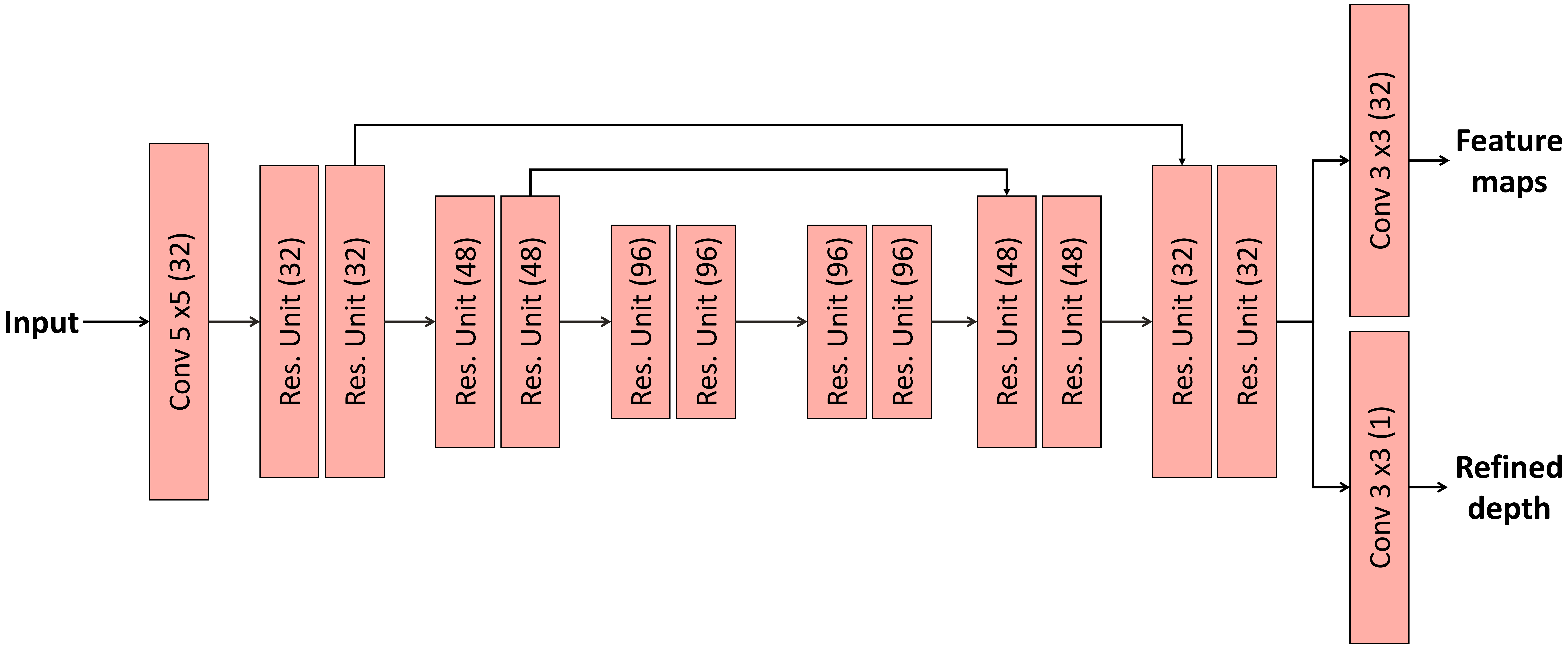}
    \caption{\textbf{Architecture  of feature extractor and depth refiner.} }
    \label{fig:unet}
\end{figure*}

%% file: tables/supp/sp_conv.tex
\begin{table*}[h]
\centering
\scalebox{0.95}{
\begin{tabular}{l|l|l}
\hline
& Layer Description & Output Dim.  \\ \hline \hline
& Input volume & D' $\times$ H' $\times$ W' $\times$ 32 \\ \hline
1-2 & ($3 \times 3 \times 3$ conv, 32 features, stride 1) $\times$ 2 
&  D' $\times$ H' $\times$ W' $\times$ 32 \\
3 & ($3 \times 3 \times 3$ conv, 32 features, stride 2) 
&  D'/2 $\times$ H'/2 $\times$ W'/2 $\times$ 32 \\
4-5 & ($3 \times 3 \times 3$ conv, 32 features, stride 1) $\times$ 2
&  D'/2 $\times$ H'/2 $\times$ W'/2 $\times$ 32 \\
6 & ($3 \times 3 \times 3$ conv, 32 features, stride 2) 
&  D'/4 $\times$ H'/4 $\times$ W'/4 $\times$ 64 \\
7-9 & ($3 \times 3 \times 3$ conv, 64 features, stride 1) $\times$ 3
&  D'/4 $\times$ H'/4 $\times$ W'/4 $\times$ 64 \\
10 & ($3 \times 3 \times 3$ conv, 64 features, stride 2) 
&  D'/8 $\times$ H'/8 $\times$ W'/8 $\times$ 128 \\
11-13 & ($3 \times 3 \times 3$ conv, 128 features, stride 1) $\times$ 3
&  D'/8 $\times$ H'/8 $\times$ W'/8 $\times$ 128 \\
14 & ($3 \times 3 \times 3$ conv, 128 features, stride 2) 
&  D'/16 $\times$ H'/16 $\times$ W'/16 $\times$ 128 \\
15-17 & ($3 \times 3 \times 3$ conv, 128 features, stride 1) $\times$ 3
&  D'/16 $\times$ H'/16 $\times$ W'/16 $\times$ 128 \\
\hline
& Resize \& Concat. outputs of layer 5, 9, 13, and 17
&  D'/16 $\times$ H'/16 $\times$ W'/16 $\times$ 352 \\

\hline

\end{tabular}
} 

\vspace{3mm}

\caption{\textbf{Architecture of SparseConvNet.} Each layer consists of sparse convolution, batch normalization and ReLU.}
\label{tab:sparse_conv}
%\vspace{-5mm}
\end{table*}

%% file: figs/gs.tex
\begin{figure}[h]
    \centering
    \includegraphics[width=\linewidth]{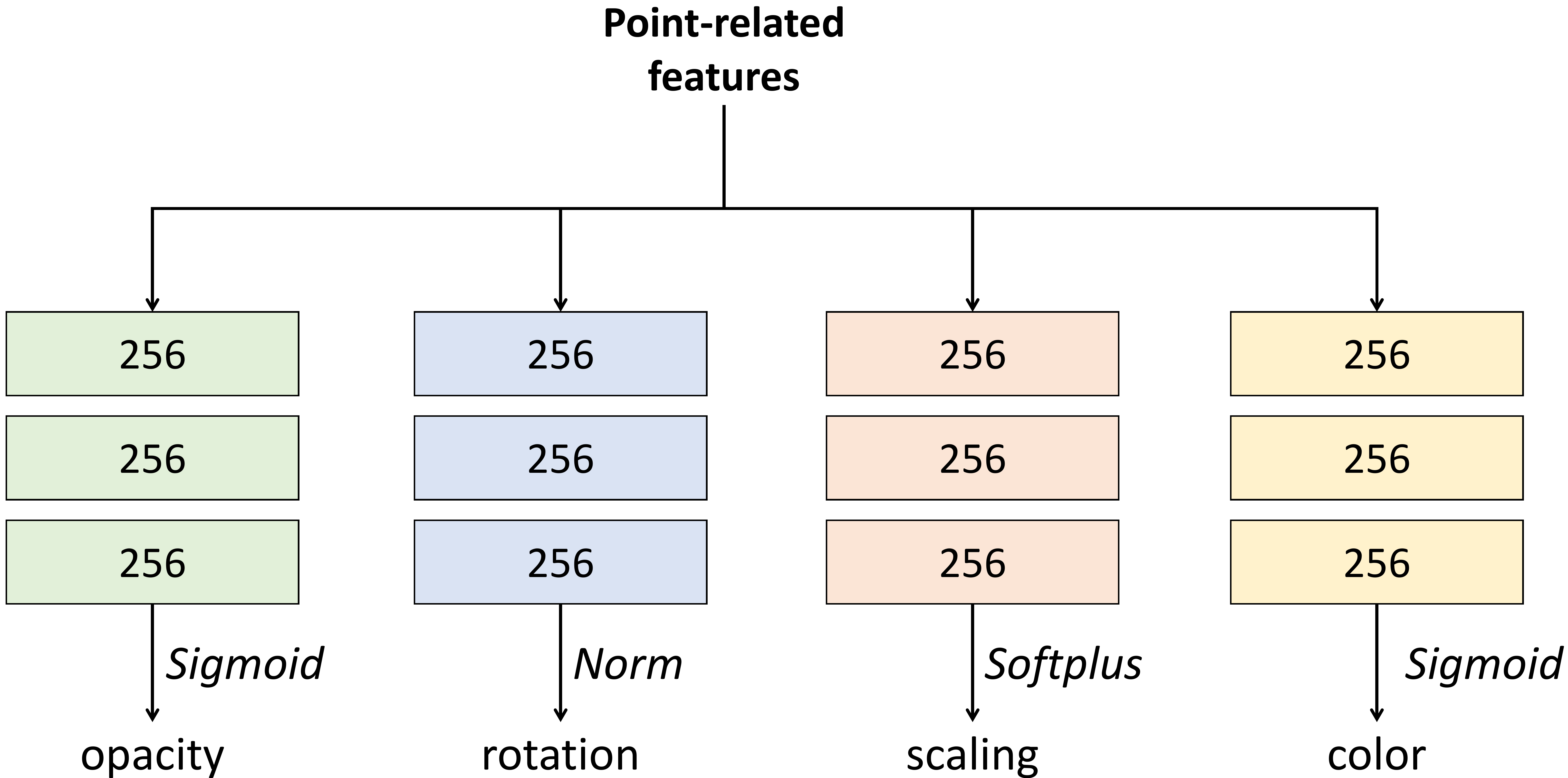}
    \caption{\textbf{Architecture  of Gaussian predictor.} }
    \label{fig:gs}
\end{figure}

%% file: figs/offset.tex
\begin{figure}[h]
    \centering
    \includegraphics[width=0.9\linewidth]{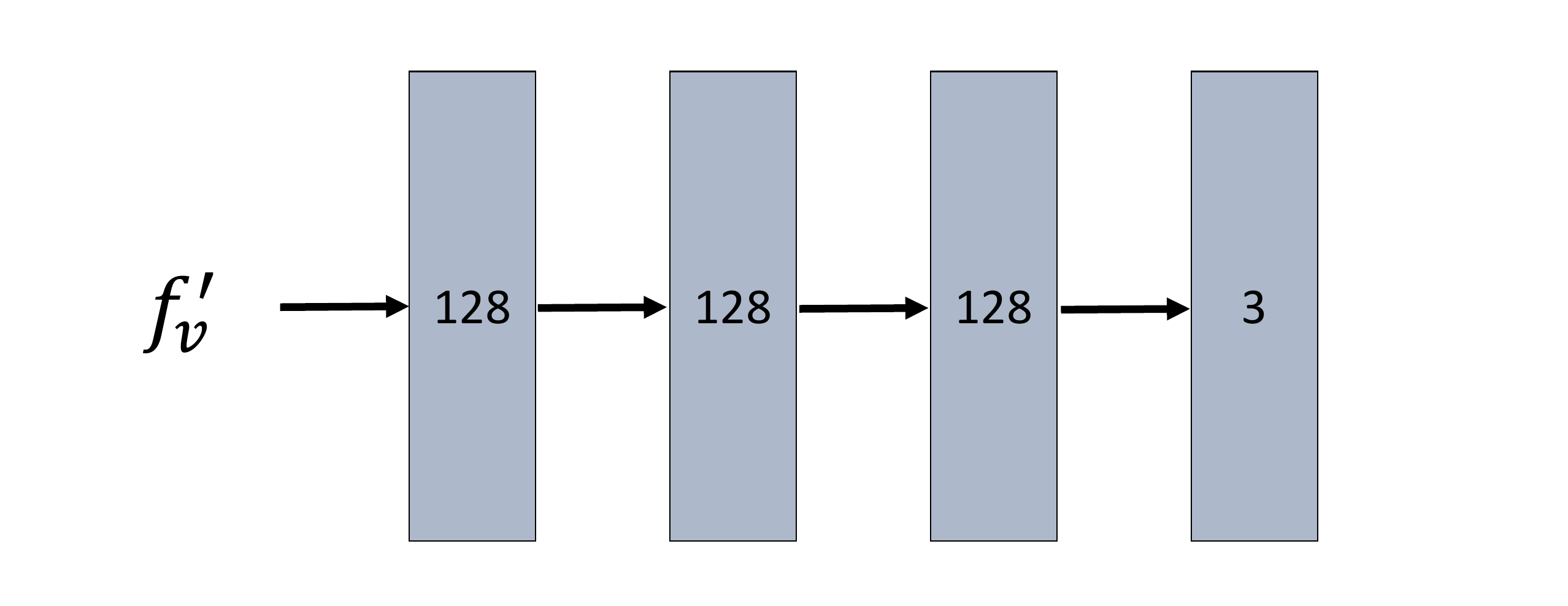}
    \caption{\textbf{Architecture  of offset estimator.} }
    \label{fig:offset}
\end{figure}

%% file: figs/thuman.tex
\begin{figure*}[h]
    \centering
    \includegraphics[width=\linewidth]{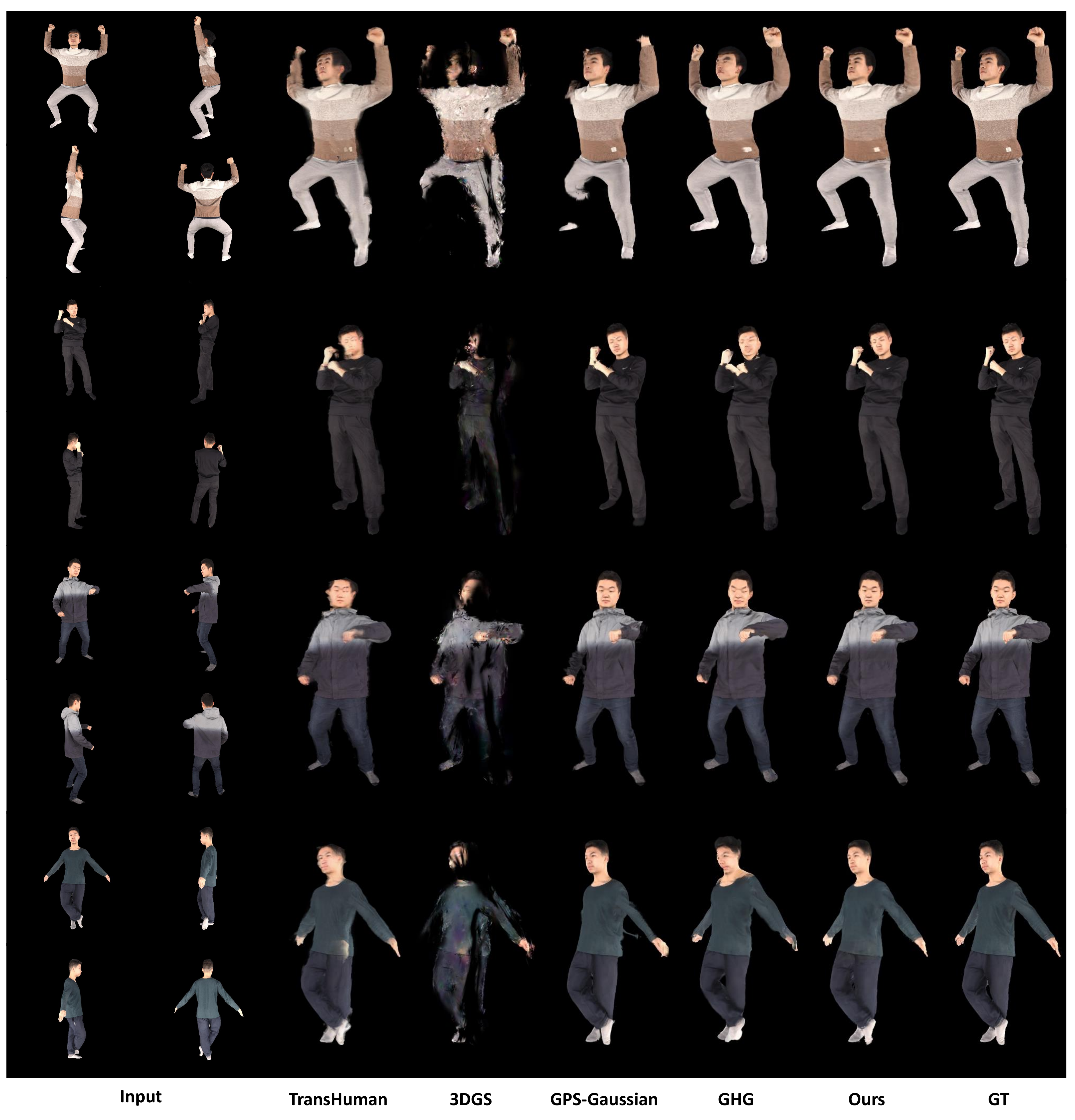}
    \caption{\textbf{More qualitative comparison of in-domain generalization.} }
    \label{fig:more_thu}
\end{figure*}

%% file: figs/cross.tex
\begin{figure*}[h]
    \centering
    \includegraphics[width=\linewidth]{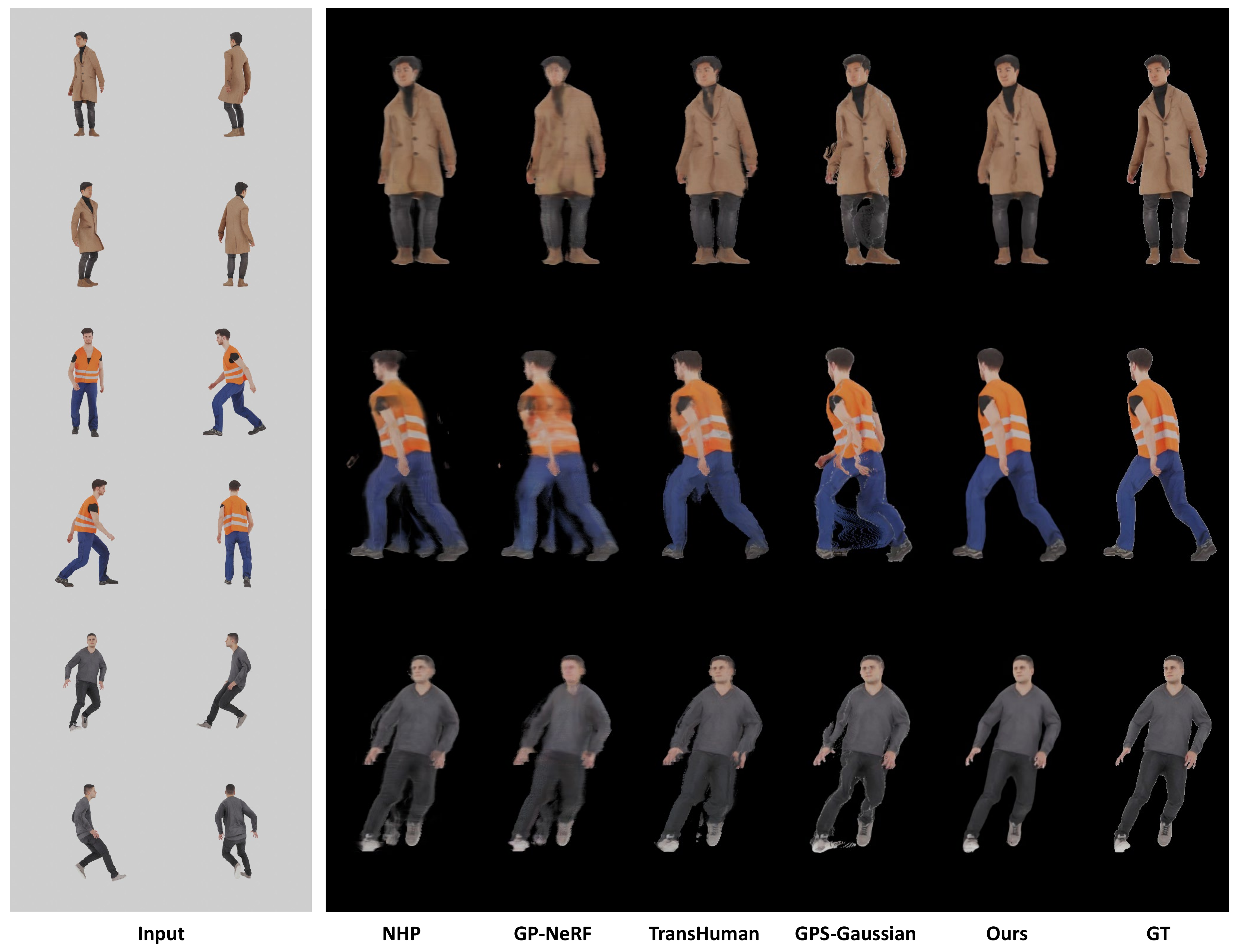}
    \caption{\textbf{More qualitative comparison of cross-domain generalization.} }
    \label{fig:more_cross}
\end{figure*}

%% file: tables/rebuttal/geo_ablation.tex
\begin{table}[]
\centering
% \vspace{-3mm}
\resizebox{\linewidth}{!}
{
\begin{tabular}{l c c c}
\toprule[1pt]
\multirow{1}{*}{Method}& \multicolumn{1}{c}{PSNR$\uparrow$}& \multicolumn{1}{c}{SSIM $\uparrow$}& \multicolumn{1}{c}{ LPIPS $\downarrow$}\\ 
\midrule
w/o depth refiner &27.27 & 0.9524&0.0602\\
w/o coarse Gaussian predictor&26.34 &0.9471 &0.0678\\
Ours Full&\textbf{28.94} &\textbf{0.9615} &\textbf{0.0433}\\
\bottomrule[1pt]
\end{tabular}
}
\caption{\textbf{More geometry-related ablation studies.}}
% \vspace{-4mm}
\label{table:geo_ablation}
\end{table}

% w/o semantic loss  & & &\\
% w/o depth refinement loss  & & &\\

%% file: tables/rebuttal/rp.tex
\begin{table}[h]
\centering
% \vspace{-3mm}
\resizebox{\linewidth}{!}
{
\begin{tabular}{l c c c}
\toprule[1pt]
\multirow{1}{*}{Method}& \multicolumn{1}{c}{PSNR$\uparrow$}& \multicolumn{1}{c}{SSIM $\uparrow$}& \multicolumn{1}{c}{ LPIPS $\downarrow$}\\ 
\midrule
NHP&26.01 &0.9384 &0.0726\\
GP-NeRF&25.33& 0.9326& 0.0792\\
TransHuman& 26.37& 0.9451& 0.0579\\
\midrule
w/o pixel-level features &26.42 & 0.9522&0.0546\\
w/o coarse-to-fine Gaussian prediction &26.58 &0.9519 &0.0594\\
Ours&\textbf{27.00} &\textbf{0.9530} &\textbf{0.0519}\\
\bottomrule[1pt]
\end{tabular}
}
% \vspace{-2mm}
\caption{\textbf{Results on RenderPeople dataset.}}
\label{table:rp}
\end{table}

%% file: main.bbl
\begin{thebibliography}{94}
\providecommand{\natexlab}[1]{#1}
\providecommand{\url}[1]{\texttt{#1}}
\expandafter\ifx\csname urlstyle\endcsname\relax
  \providecommand{\doi}[1]{doi: #1}\else
  \providecommand{\doi}{doi: \begingroup \urlstyle{rm}\Url}\fi

\bibitem[ren()]{renderpeople}
Renderpeople.
\newblock \url{https://renderpeople.com/}.

\bibitem[eas(2021)]{easymocap}
Easymocap - make human motion capture easier.
\newblock Github, 2021.

\bibitem[Aliev et~al.(2020)Aliev, Sevastopolsky, Kolos, Ulyanov, and Lempitsky]{aliev2020neural}
Kara-Ali Aliev, Artem Sevastopolsky, Maria Kolos, Dmitry Ulyanov, and Victor Lempitsky.
\newblock Neural point-based graphics.
\newblock In \emph{ECCV}, 2020.

\bibitem[Charatan et~al.(2024)Charatan, Li, Tagliasacchi, and Sitzmann]{charatan23pixelsplat}
David Charatan, Sizhe Li, Andrea Tagliasacchi, and Vincent Sitzmann.
\newblock {pixelSplat}: 3d gaussian splats from image pairs for scalable generalizable 3d reconstruction.
\newblock In \emph{CVPR}, 2024.

\bibitem[Chen et~al.(2021)Chen, Xu, Zhao, Zhang, Xiang, Yu, and Su]{chen2021mvsnerf}
Anpei Chen, Zexiang Xu, Fuqiang Zhao, Xiaoshuai Zhang, Fanbo Xiang, Jingyi Yu, and Hao Su.
\newblock {MVSNeRF}: Fast generalizable radiance field reconstruction from multi-view stereo.
\newblock In \emph{ICCV}, 2021.

\bibitem[Chen et~al.(2022{\natexlab{a}})Chen, Xu, Geiger, Yu, and Su]{Chen2022ECCV}
Anpei Chen, Zexiang Xu, Andreas Geiger, Jingyi Yu, and Hao Su.
\newblock {TensoRF}: Tensorial radiance fields.
\newblock In \emph{ECCV}, 2022{\natexlab{a}}.

\bibitem[Chen et~al.(2024{\natexlab{a}})Chen, Li, Zhang, Chen, Huang, and Lee]{chen2024generalizable}
Jinnan Chen, Chen Li, Jianfeng Zhang, Hanlin Chen, Buzhen Huang, and Gim~Hee Lee.
\newblock Generalizable human gaussians from single-view image, 2024{\natexlab{a}}.

\bibitem[Chen et~al.(2024{\natexlab{b}})Chen, Li, Zhang, Zhu, Huang, Chen, , and Lee]{jnchen24hgm}
Jinnan Chen, Chen Li, Jianfeng Zhang, Lingting Zhu, Buzhen Huang, Hanlin Chen, , and Gim~Hee Lee.
\newblock Generalizable human gaussians from single-view image.
\newblock 2024{\natexlab{b}}.

\bibitem[Chen et~al.(2022{\natexlab{b}})Chen, Zhang, Xu, Liu, Cai, Feng, and Yan]{chen2022gpnerf}
Mingfei Chen, Jianfeng Zhang, Xiangyu Xu, Lijuan Liu, Yujun Cai, Jiashi Feng, and Shuicheng Yan.
\newblock Geometry-guided progressive nerf for generalizable and efficient neural human rendering.
\newblock In \emph{ECCV}, 2022{\natexlab{b}}.

\bibitem[Chen et~al.(2023)Chen, Wang, Chen, Zhang, Li, Guo, Wang, and Wang]{Chen_2023_CVPR}
Yue Chen, Xuan Wang, Xingyu Chen, Qi Zhang, Xiaoyu Li, Yu Guo, Jue Wang, and Fei Wang.
\newblock Uv volumes for real-time rendering of editable free-view human performance.
\newblock In \emph{CVPR}, 2023.

\bibitem[Chen et~al.(2024{\natexlab{c}})Chen, Xu, Zheng, Zhuang, Pollefeys, Geiger, Cham, and Cai]{chen2024mvsplat}
Yuedong Chen, Haofei Xu, Chuanxia Zheng, Bohan Zhuang, Marc Pollefeys, Andreas Geiger, Tat-Jen Cham, and Jianfei Cai.
\newblock {MVSplat}: Efficient 3d gaussian splatting from sparse multi-view images.
\newblock In \emph{ECCV}, 2024{\natexlab{c}}.

\bibitem[Chen et~al.(2024{\natexlab{d}})Chen, Zheng, Li, Xu, and Liu]{chen2024meshavatar}
Yushuo Chen, Zerong Zheng, Zhe Li, Chao Xu, and Yebin Liu.
\newblock {MeshAvatar}: Learning high-quality triangular human avatars from multi-view videos.
\newblock In \emph{ECCV}, 2024{\natexlab{d}}.

\bibitem[Chen and Liu(2022)]{chen2022relighting}
Zhaoxi Chen and Ziwei Liu.
\newblock {Relighting4D}: Neural relightable human from videos.
\newblock In \emph{ECCV}, 2022.

\bibitem[Chen et~al.(2024{\natexlab{e}})Chen, Yang, Huang, de~Lutio, Esturo, Ivanovic, Litany, Gojcic, Fidler, Pavone, Song, and Wang]{chen2024omnire}
Ziyu Chen, Jiawei Yang, Jiahui Huang, Riccardo de Lutio, Janick~Martinez Esturo, Boris Ivanovic, Or Litany, Zan Gojcic, Sanja Fidler, Marco Pavone, Li Song, and Yue Wang.
\newblock {OmniRe}: Omni urban scene reconstruction.
\newblock \emph{arXiv preprint arXiv:2408.16760}, 2024{\natexlab{e}}.

\bibitem[Chu and Harada(2024)]{chu2024gagavatar}
Xuangeng Chu and Tatsuya Harada.
\newblock Generalizable and animatable gaussian head avatar.
\newblock In \emph{NeurIPS}, 2024.

\bibitem[Fang et~al.(2022)Fang, Yi, Wang, Xie, Zhang, Liu, Nie\ss{}ner, and Tian]{TiNeuVox}
Jiemin Fang, Taoran Yi, Xinggang Wang, Lingxi Xie, Xiaopeng Zhang, Wenyu Liu, Matthias Nie\ss{}ner, and Qi Tian.
\newblock Fast dynamic radiance fields with time-aware neural voxels.
\newblock In \emph{SIGGRAPH Asia}, 2022.

\bibitem[Fridovich-Keil et~al.(2022)Fridovich-Keil, Yu, Tancik, Chen, Recht, and Kanazawa]{yu2022plenoxels}
Sara Fridovich-Keil, Alex Yu, Matthew Tancik, Qinhong Chen, Benjamin Recht, and Angjoo Kanazawa.
\newblock {Plenoxels}: Radiance fields without neural networks.
\newblock In \emph{CVPR}, 2022.

\bibitem[Gao et~al.(2023)Gao, Wang, Liu, Liu, Theobalt, and Chen]{gao2023neural}
Qingzhe Gao, Yiming Wang, Libin Liu, Lingjie Liu, Christian Theobalt, and Baoquan Chen.
\newblock {Neural novel actor}: Learning a generalized animatable neural representation for human actors.
\newblock \emph{IEEE Transactions on Visualization and Computer Graphics}, 2023.

\bibitem[Gao et~al.(2022)Gao, Yang, Kim, Peng, Liu, and Tong]{Gao_2022_mpsnerf}
Xiangjun Gao, Jiaolong Yang, Jongyoo Kim, Sida Peng, Zicheng Liu, and Xin Tong.
\newblock {MPS-NeRF}: Generalizable 3d human rendering from multiview images.
\newblock \emph{IEEE Transactions on Pattern Analysis and Machine Intelligence}, pages 1--12, 2022.

\bibitem[Graham et~al.(2018)Graham, Engelcke, and van~der Maaten]{Graham_2018_CVPR}
Benjamin Graham, Martin Engelcke, and Laurens van~der Maaten.
\newblock 3d semantic segmentation with submanifold sparse convolutional networks.
\newblock In \emph{CVPR}, 2018.

\bibitem[Guo et~al.(2023)Guo, Jiang, Chen, Song, and Hilliges]{guo2023vid2avatar}
Chen Guo, Tianjian Jiang, Xu Chen, Jie Song, and Otmar Hilliges.
\newblock {Vid2avatar}: 3d avatar reconstruction from videos in the wild via self-supervised scene decomposition.
\newblock In \emph{CVPR}, 2023.

\bibitem[Hu et~al.(2024)Hu, Zhang, Zhang, Zhou, Liu, Zhang, and Nie]{hu2024gaussianavatar}
Liangxiao Hu, Hongwen Zhang, Yuxiang Zhang, Boyao Zhou, Boning Liu, Shengping Zhang, and Liqiang Nie.
\newblock {GaussianAvatar}: Towards realistic human avatar modeling from a single video via animatable 3d gaussians.
\newblock In \emph{CVPR}, 2024.

\bibitem[Hu and Liu(2024)]{hu2024gauhuman}
Shoukang Hu and Ziwei Liu.
\newblock {GauHuman}: Articulated gaussian splatting from monocular human videos.
\newblock In \emph{CVPR}, 2024.

\bibitem[Hu et~al.(2023{\natexlab{a}})Hu, Hong, Pan, Mei, Yang, and Liu]{hu2023sherf}
Shoukang Hu, Fangzhou Hong, Liang Pan, Haiyi Mei, Lei Yang, and Ziwei Liu.
\newblock {SHERF}: Generalizable human nerf from a single image.
\newblock \emph{ICCV}, 2023{\natexlab{a}}.

\bibitem[Hu et~al.(2023{\natexlab{b}})Hu, Xu, Liu, and Jia]{Hu_2023_CVPR}
Tao Hu, Xiaogang Xu, Shu Liu, and Jiaya Jia.
\newblock {Point2Pix}: Photo-realistic point cloud rendering via neural radiance fields.
\newblock In \emph{CVPR}, 2023{\natexlab{b}}.

\bibitem[Huang et~al.(2024)Huang, Sun, Yang, Lyu, Cao, and Qi]{huang2023sc}
Yi-Hua Huang, Yang-Tian Sun, Ziyi Yang, Xiaoyang Lyu, Yan-Pei Cao, and Xiaojuan Qi.
\newblock {SC-GS}: Sparse-controlled gaussian splatting for editable dynamic scenes.
\newblock In \emph{CVPR}, 2024.

\bibitem[Jiang et~al.(2022)Jiang, Yi, Samei, Tuzel, and Ranjan]{jiang2022neuman}
Wei Jiang, Kwang~Moo Yi, Golnoosh Samei, Oncel Tuzel, and Anurag Ranjan.
\newblock {Neuman}: Neural human radiance field from a single video.
\newblock In \emph{ECCV}, 2022.

\bibitem[Jiang et~al.(2024{\natexlab{a}})Jiang, Shen, Hong, Guo, Wu, Zhang, Yu, and Xu]{jiang2024dualgs}
Yuheng Jiang, Zhehao Shen, Yu Hong, Chengcheng Guo, Yize Wu, Yingliang Zhang, Jingyi Yu, and Lan Xu.
\newblock Robust dual gaussian splatting for immersive human-centric volumetric videos.
\newblock In \emph{Siggraph Asia}, 2024{\natexlab{a}}.

\bibitem[Jiang et~al.(2024{\natexlab{b}})Jiang, Shen, Wang, Su, Hong, Zhang, Yu, and Xu]{jiang2024hifi4g}
Yuheng Jiang, Zhehao Shen, Penghao Wang, Zhuo Su, Yu Hong, Yingliang Zhang, Jingyi Yu, and Lan Xu.
\newblock {HiFi4G}: High-fidelity human performance rendering via compact gaussian splatting.
\newblock In \emph{CVPR}, 2024{\natexlab{b}}.

\bibitem[Jiaxu et~al.()Jiaxu, Zhang, and Xu]{jiaxulearning}
WANG Jiaxu, Ziyi Zhang, and Renjing Xu.
\newblock Learning robust generalizable radiance field with visibility and feature augmented point representation.
\newblock In \emph{ICLR}.

\bibitem[Kerbl et~al.(2023)Kerbl, Kopanas, Leimk{\"u}hler, and Drettakis]{3DGS}
Bernhard Kerbl, Georgios Kopanas, Thomas Leimk{\"u}hler, and George Drettakis.
\newblock 3d gaussian splatting for real-time radiance field rendering.
\newblock \emph{ACM Transactions on Graphics}, 42\penalty0 (4), 2023.

\bibitem[Kocabas et~al.(2024)Kocabas, Chang, Gabriel, Tuzel, and Ranjan]{kocabas2024hugs}
Muhammed Kocabas, Jen-Hao~Rick Chang, James Gabriel, Oncel Tuzel, and Anurag Ranjan.
\newblock {HUGS}: Human gaussian splatting.
\newblock In \emph{CVPR}, 2024.

\bibitem[Kopanas et~al.(2022)Kopanas, Leimk{\"u}hler, Rainer, Jambon, and Drettakis]{kopanas2022neural}
Georgios Kopanas, Thomas Leimk{\"u}hler, Gilles Rainer, Cl{\'e}ment Jambon, and George Drettakis.
\newblock Neural point catacaustics for novel-view synthesis of reflections.
\newblock \emph{ACM Transactions on Graphics (TOG)}, 41\penalty0 (6):\penalty0 1--15, 2022.

\bibitem[Kwon et~al.(2021)Kwon, Kim, Ceylan, and Fuchs]{kwon2021neural}
Youngjoong Kwon, Dahun Kim, Duygu Ceylan, and Henry Fuchs.
\newblock {Neural human performer}: Learning generalizable radiance fields for human performance rendering.
\newblock In \emph{NeurIPS}, 2021.

\bibitem[Kwon et~al.(2023)Kwon, Kim, Ceylan, and Fuchs]{kwon2023nip}
Youngjoong Kwon, Dahun Kim, Duygu Ceylan, and Henry Fuchs.
\newblock {Neural Image-based Avatars}: Generalizable radiance fields for human avatar modeling.
\newblock In \emph{ICLR}, 2023.

\bibitem[Kwon et~al.(2024)Kwon, Fang, Lu, Dong, Zhang, Carrasco, Mosella-Montoro, Xu, Takagi, Kim, Prakash, and la~Torre]{kwon2024ghg}
Youngjoong Kwon, Baole Fang, Yixing Lu, Haoye Dong, Cheng Zhang, Francisco~Vicente Carrasco, Albert Mosella-Montoro, Jianjin Xu, Shingo Takagi, Daeil Kim, Aayush Prakash, and Fernando~De la Torre.
\newblock Generalizable human gaussians for sparse view synthesis.
\newblock In \emph{ECCV}, 2024.

\bibitem[Lassner and Zollhofer(2021)]{lassner2021pulsar}
Christoph Lassner and Michael Zollhofer.
\newblock {Pulsar}: Efficient sphere-based neural rendering.
\newblock In \emph{CVPR}, 2021.

\bibitem[Li et~al.(2022)Li, Tanke, Vo, Zollhofer, Gall, Kanazawa, and Lassner]{li2022tava}
Ruilong Li, Julian Tanke, Minh Vo, Michael Zollhofer, Jurgen Gall, Angjoo Kanazawa, and Christoph Lassner.
\newblock {TAVA}: Template-free animatable volumetric actors.
\newblock In \emph{ECCV}, 2022.

\bibitem[Li et~al.(2023)Li, Zheng, Liu, Zhou, and Liu]{li2023posevocab}
Zhe Li, Zerong Zheng, Yuxiao Liu, Boyao Zhou, and Yebin Liu.
\newblock {PoseVocab}: Learning joint-structured pose embeddings for human avatar modeling.
\newblock In \emph{SIGGRAPH}, 2023.

\bibitem[Li et~al.(2024)Li, Zheng, Wang, and Liu]{li2024animatablegaussians}
Zhe Li, Zerong Zheng, Lizhen Wang, and Yebin Liu.
\newblock {Animatable Gaussians}: Learning pose-dependent gaussian maps for high-fidelity human avatar modeling.
\newblock In \emph{CVPR}, 2024.

\bibitem[Lin et~al.(2022{\natexlab{a}})Lin, Peng, Xu, Yan, Shuai, Bao, and Zhou]{lin2022enerf}
Haotong Lin, Sida Peng, Zhen Xu, Yunzhi Yan, Qing Shuai, Hujun Bao, and Xiaowei Zhou.
\newblock Efficient neural radiance fields for interactive free-viewpoint video.
\newblock In \emph{SIGGRAPH Asia}, 2022{\natexlab{a}}.

\bibitem[Lin et~al.(2022{\natexlab{b}})Lin, Zhang, Zheng, Shao, and Liu]{lin2022learning}
Siyou Lin, Hongwen Zhang, Zerong Zheng, Ruizhi Shao, and Yebin Liu.
\newblock Learning implicit templates for point-based clothed human modeling.
\newblock In \emph{ECCV}, 2022{\natexlab{b}}.

\bibitem[Liu et~al.(2015)Liu, Wang, Foroosh, Tappen, and Pensky]{Liu_2015_CVPR}
Baoyuan Liu, Min Wang, Hassan Foroosh, Marshall Tappen, and Marianna Pensky.
\newblock Sparse convolutional neural networks.
\newblock In \emph{CVPR}, 2015.

\bibitem[Liu et~al.(2020)Liu, Gu, Zaw~Lin, Chua, and Theobalt]{Liu2020NSVF}
Lingjie Liu, Jiatao Gu, Kyaw Zaw~Lin, Tat-Seng Chua, and Christian Theobalt.
\newblock In \emph{NeurIPS}, 2020.

\bibitem[Liu et~al.(2021)Liu, Habermann, Rudnev, Sarkar, Gu, and Theobalt]{liu2021neural}
Lingjie Liu, Marc Habermann, Viktor Rudnev, Kripasindhu Sarkar, Jiatao Gu, and Christian Theobalt.
\newblock {Neural Actor}: Neural free-view synthesis of human actors with pose control.
\newblock \emph{ACM Trans. Graphics}, 40\penalty0 (6), 2021.

\bibitem[Liu et~al.(2025)Liu, Wang, Hu, Shen, Ye, Zang, Cao, Li, and Liu]{liu2025mvsgaussian}
Tianqi Liu, Guangcong Wang, Shoukang Hu, Liao Shen, Xinyi Ye, Yuhang Zang, Zhiguo Cao, Wei Li, and Ziwei Liu.
\newblock {MVSGaussian}: Fast generalizable gaussian splatting reconstruction from multi-view stereo.
\newblock In \emph{ECCV}, pages 37--53. Springer, 2025.

\bibitem[Loper et~al.(2015)Loper, Mahmood, Romero, Pons-Moll, and Black]{SMPL:2015}
Matthew Loper, Naureen Mahmood, Javier Romero, Gerard Pons-Moll, and Michael~J. Black.
\newblock {SMPL}: A skinned multi-person linear model.
\newblock \emph{ACM Trans. Graphics}, 34\penalty0 (6):\penalty0 248:1--248:16, 2015.

\bibitem[Luiten et~al.(2024)Luiten, Kopanas, Leibe, and Ramanan]{luiten2023dynamic}
Jonathon Luiten, Georgios Kopanas, Bastian Leibe, and Deva Ramanan.
\newblock {Dynamic 3D Gaussians}: Tracking by persistent dynamic view synthesis.
\newblock In \emph{3DV}, 2024.

\bibitem[Mihajlovic et~al.(2022)Mihajlovic, Bansal, Zollhoefer, Tang, and Saito]{mihajlovic2022keypointnerf}
Marko Mihajlovic, Aayush Bansal, Michael Zollhoefer, Siyu Tang, and Shunsuke Saito.
\newblock {KeypointNeRF}: Generalizing image-based volumetric avatars using relative spatial encoding of keypoints.
\newblock In \emph{ECCV}, 2022.

\bibitem[Mildenhall et~al.(2020)Mildenhall, Srinivasan, Tancik, Barron, Ramamoorthi, and Ng]{mildenhall2020nerf}
Ben Mildenhall, Pratul~P. Srinivasan, Matthew Tancik, Jonathan~T. Barron, Ravi Ramamoorthi, and Ren Ng.
\newblock {NeRF}: Representing scenes as neural radiance fields for view synthesis.
\newblock In \emph{ECCV}, 2020.

\bibitem[Moon et~al.(2024)Moon, Shiratori, and Saito]{moon2024exavatar}
Gyeongsik Moon, Takaaki Shiratori, and Shunsuke Saito.
\newblock Expressive whole-body 3d gaussian avatar.
\newblock In \emph{ECCV}, 2024.

\bibitem[Moreau et~al.(2024)Moreau, Song, Dhamo, Shaw, Zhou, and P{\'e}rez-Pellitero]{moreau2024human}
Arthur Moreau, Jifei Song, Helisa Dhamo, Richard Shaw, Yiren Zhou, and Eduardo P{\'e}rez-Pellitero.
\newblock {Human gaussian splatting}: Real-time rendering of animatable avatars.
\newblock In \emph{CVPR}, 2024.

\bibitem[M\"uller et~al.(2022)M\"uller, Evans, Schied, and Keller]{mueller2022instant}
Thomas M\"uller, Alex Evans, Christoph Schied, and Alexander Keller.
\newblock Instant neural graphics primitives with a multiresolution hash encoding.
\newblock \emph{ACM Trans. Graph.}, 41\penalty0 (4):\penalty0 102:1--102:15, 2022.

\bibitem[Oquab et~al.(2023)Oquab, Darcet, Moutakanni, Vo, Szafraniec, Khalidov, Fernandez, Haziza, Massa, El-Nouby, Howes, Huang, Xu, Sharma, Li, Galuba, Rabbat, Assran, Ballas, Synnaeve, Misra, Jegou, Mairal, Labatut, Joulin, and Bojanowski]{oquab2023dinov2}
Maxime Oquab, Timothée Darcet, Theo Moutakanni, Huy~V. Vo, Marc Szafraniec, Vasil Khalidov, Pierre Fernandez, Daniel Haziza, Francisco Massa, Alaaeldin El-Nouby, Russell Howes, Po-Yao Huang, Hu Xu, Vasu Sharma, Shang-Wen Li, Wojciech Galuba, Mike Rabbat, Mido Assran, Nicolas Ballas, Gabriel Synnaeve, Ishan Misra, Herve Jegou, Julien Mairal, Patrick Labatut, Armand Joulin, and Piotr Bojanowski.
\newblock {DINOv2}: Learning robust visual features without supervision.
\newblock \emph{TMLR}, 2023.

\bibitem[Pan et~al.(2024)Pan, Su, Lin, Fan, Zhang, Li, Shen, Mu, and Liu]{pan2024humansplat}
Panwang Pan, Zhuo Su, Chenguo Lin, Zhen Fan, Yongjie Zhang, Zeming Li, Tingting Shen, Yadong Mu, and Yebin Liu.
\newblock {HumanSplat}: Generalizable single-image human gaussian splatting with structure priors.
\newblock In \emph{NeurIPS}, 2024.

\bibitem[Pan et~al.(2023)Pan, Yang, Ma, Zhou, and Yang]{Pan_2023_ICCV}
Xiao Pan, Zongxin Yang, Jianxin Ma, Chang Zhou, and Yi Yang.
\newblock {TransHuman}: A transformer-based human representation for generalizable neural human rendering.
\newblock In \emph{ICCV}, 2023.

\bibitem[Peng et~al.(2020)Peng, Niemeyer, Mescheder, Pollefeys, and Geiger]{Peng2020ECCV}
Songyou Peng, Michael Niemeyer, Lars Mescheder, Marc Pollefeys, and Andreas Geiger.
\newblock Convolutional occupancy networks.
\newblock In \emph{ECCV}, 2020.

\bibitem[Peng et~al.(2021)Peng, Zhang, Xu, Wang, Shuai, Bao, and Zhou]{peng2021neural}
Sida Peng, Yuanqing Zhang, Yinghao Xu, Qianqian Wang, Qing Shuai, Hujun Bao, and Xiaowei Zhou.
\newblock {Neural Body}: Implicit neural representations with structured latent codes for novel view synthesis of dynamic humans.
\newblock In \emph{CVPR}, 2021.

\bibitem[Prospero et~al.(2024)Prospero, Hamdi, Henriques, and Rupprecht]{prospero2024gstprecise3dhuman}
Lorenza Prospero, Abdullah Hamdi, Joao~F. Henriques, and Christian Rupprecht.
\newblock {GST}: Precise 3d human body from a single image with gaussian splatting transformers, 2024.

\bibitem[R{\"u}ckert et~al.(2022)R{\"u}ckert, Franke, and Stamminger]{ruckert2022adop}
Darius R{\"u}ckert, Linus Franke, and Marc Stamminger.
\newblock {Adop}: Approximate differentiable one-pixel point rendering.
\newblock \emph{ACM Transactions on Graphics (ToG)}, 41\penalty0 (4):\penalty0 1--14, 2022.

\bibitem[Saito et~al.(2019)Saito, Huang, Natsume, Morishima, Kanazawa, and Li]{pifuSHNMKL19}
Shunsuke Saito, Zeng Huang, Ryota Natsume, Shigeo Morishima, Angjoo Kanazawa, and Hao Li.
\newblock {PIFu}: Pixel-aligned implicit function for high-resolution clothed human digitization.
\newblock In \emph{ICCV}, 2019.

\bibitem[{Sara Fridovich-Keil and Giacomo Meanti} et~al.(2023){Sara Fridovich-Keil and Giacomo Meanti}, Warburg, Recht, and Kanazawa]{kplanes_2023}
{Sara Fridovich-Keil and Giacomo Meanti}, Frederik~Rahbæk Warburg, Benjamin Recht, and Angjoo Kanazawa.
\newblock {K-Planes}: Explicit radiance fields in space, time, and appearance.
\newblock In \emph{CVPR}, 2023.

\bibitem[Shao et~al.(2023)Shao, Zheng, Tu, Liu, Zhang, and Liu]{shao2023tensor4d}
Ruizhi Shao, Zerong Zheng, Hanzhang Tu, Boning Liu, Hongwen Zhang, and Yebin Liu.
\newblock {Tensor4d}: Efficient neural 4d decomposition for high-fidelity dynamic reconstruction and rendering.
\newblock In \emph{CVPR}, 2023.

\bibitem[Shi et~al.(2020)Shi, Guo, Jiang, Wang, Shi, Wang, and Li]{shi2020pv}
Shaoshuai Shi, Chaoxu Guo, Li Jiang, Zhe Wang, Jianping Shi, Xiaogang Wang, and Hongsheng Li.
\newblock {PV-RCNN}: Point-voxel feature set abstraction for 3d object detection.
\newblock In \emph{CVPR}, 2020.

\bibitem[Szymanowicz et~al.(2024)Szymanowicz, Rupprecht, and Vedaldi]{szymanowicz24splatter}
Stanislaw Szymanowicz, Christian Rupprecht, and Andrea Vedaldi.
\newblock {Splatter Image}: Ultra-fast single-view 3d reconstruction.
\newblock In \emph{CVPR}, 2024.

\bibitem[Wang et~al.(2024{\natexlab{a}})Wang, Zhang, and Xu]{wang24ICLR}
Jiaxu Wang, Ziyi Zhang, and Renjing Xu.
\newblock Learning robust generalizable radiance field with visibility and feature augmented point representation.
\newblock In \emph{ICLR}, 2024{\natexlab{a}}.

\bibitem[Wang et~al.(2024{\natexlab{b}})Wang, Zhang, Wang, Yao, Xie, Yu, Wu, and Xu]{wang2024v3}
Penghao Wang, Zhirui Zhang, Liao Wang, Kaixin Yao, Siyuan Xie, Jingyi Yu, Minye Wu, and Lan Xu.
\newblock {V3}: Viewing volumetric videos on mobiles via streamable 2d dynamic gaussians.
\newblock In \emph{Siggraph Asia}, 2024{\natexlab{b}}.

\bibitem[Wang et~al.(2021)Wang, Wang, Genova, Srinivasan, Zhou, Barron, Martin-Brualla, Snavely, and Funkhouser]{wang2021ibrnet}
Qianqian Wang, Zhicheng Wang, Kyle Genova, Pratul Srinivasan, Howard Zhou, Jonathan~T. Barron, Ricardo Martin-Brualla, Noah Snavely, and Thomas Funkhouser.
\newblock {IBRNet}: Learning multi-view image-based rendering.
\newblock In \emph{CVPR}, 2021.

\bibitem[Wang et~al.(2022)Wang, Schwarz, Geiger, and Tang]{ARAH:2022:ECCV}
Shaofei Wang, Katja Schwarz, Andreas Geiger, and Siyu Tang.
\newblock {ARAH}: Animatable volume rendering of articulated human sdfs.
\newblock In \emph{ECCV}, 2022.

\bibitem[Wang et~al.(2004)Wang, Bovik, Sheikh, and Simoncelli]{ssim}
Zhou Wang, A.C. Bovik, H.R. Sheikh, and E.P. Simoncelli.
\newblock Image quality assessment: from error visibility to structural similarity.
\newblock \emph{IEEE Transactions on Image Processing}, 13\penalty0 (4):\penalty0 600--612, 2004.

\bibitem[Weng et~al.(2022)Weng, Curless, Srinivasan, Barron, and Kemelmacher-Shlizerman]{weng2022humannerf}
Chung-Yi Weng, Brian Curless, Pratul~P Srinivasan, Jonathan~T Barron, and Ira Kemelmacher-Shlizerman.
\newblock {HumanNeRF}: Free-viewpoint rendering of moving people from monocular video.
\newblock In \emph{CVPR}, 2022.

\bibitem[Wiles et~al.(2020)Wiles, Gkioxari, Szeliski, and Johnson]{wiles2020synsin}
Olivia Wiles, Georgia Gkioxari, Richard Szeliski, and Justin Johnson.
\newblock {Synsin}: End-to-end view synthesis from a single image.
\newblock In \emph{CVPR}, 2020.

\bibitem[Wu et~al.(2024)Wu, Yi, Fang, Xie, Zhang, Wei, Liu, Tian, and Xinggang]{wu20234dgaussians}
Guanjun Wu, Taoran Yi, Jiemin Fang, Lingxi Xie, Xiaopeng Zhang, Wei Wei, Wenyu Liu, Qi Tian, and Wang Xinggang.
\newblock 4d gaussian splatting for real-time dynamic scene rendering.
\newblock In \emph{CVPR}, 2024.

\bibitem[Xiang et~al.(2021)Xiang, Wen, Liu, Cao, Wan, Zheng, and Han]{xiang2021snowflakenet}
Peng Xiang, Xin Wen, Yu-Shen Liu, Yan-Pei Cao, Pengfei Wan, Wen Zheng, and Zhizhong Han.
\newblock {SnowflakeNet}: Point cloud completion by snowflake point deconvolution with skip-transformer.
\newblock In \emph{ICCV}, 2021.

\bibitem[Xiang et~al.(2023)Xiang, Wen, Liu, Cao, Wan, Zheng, and Han]{xiang2023SPD}
Peng Xiang, Xin Wen, Yu-Shen Liu, Yan-Pei Cao, Pengfei Wan, Wen Zheng, and Zhizhong Han.
\newblock Snowflake point deconvolution for point cloud completion and generation with skip-transformer.
\newblock \emph{IEEE Transactions on Pattern Analysis and Machine Intelligence}, 45\penalty0 (5):\penalty0 6320--6338, 2023.

\bibitem[Xiao et~al.(2024)Xiao, Zhang, Xu, and Zheng]{NECA2024CVPR}
Junjin Xiao, Qing Zhang, Zhan Xu, and Wei-Shi Zheng.
\newblock {NECA}: Neural customizable human avatar.
\newblock In \emph{CVPR}, 2024.

\bibitem[Xiu et~al.(2022)Xiu, Yang, Tzionas, and Black]{xiu2022icon}
Yuliang Xiu, Jinlong Yang, Dimitrios Tzionas, and Michael~J. Black.
\newblock {ICON}: {I}mplicit {C}lothed humans {O}btained from {N}ormals.
\newblock In \emph{CVPR}, 2022.

\bibitem[Xu et~al.(2022{\natexlab{a}})Xu, Xu, Philip, Bi, Shu, Sunkavalli, and Neumann]{xu2022point}
Qiangeng Xu, Zexiang Xu, Julien Philip, Sai Bi, Zhixin Shu, Kalyan Sunkavalli, and Ulrich Neumann.
\newblock {Point-nerf}: Point-based neural radiance fields.
\newblock In \emph{CVPR}, 2022{\natexlab{a}}.

\bibitem[Xu et~al.(2022{\natexlab{b}})Xu, Fujita, and Matsumoto]{Xu_2022_CVPR}
Tianhan Xu, Yasuhiro Fujita, and Eiichi Matsumoto.
\newblock Surface-aligned neural radiance fields for controllable 3d human synthesis.
\newblock In \emph{CVPR}, 2022{\natexlab{b}}.

\bibitem[Xu et~al.(2024{\natexlab{a}})Xu, Wang, Zheng, Su, and Liu]{xu2023gphm}
Yuelang Xu, Lizhen Wang, Zerong Zheng, Zhaoqi Su, and Yebin Liu.
\newblock 3d gaussian parametric head model.
\newblock In \emph{ECCV}, 2024{\natexlab{a}}.

\bibitem[Xu et~al.(2024{\natexlab{b}})Xu, Peng, Geng, Mou, Yan, Sun, Bao, and Zhou]{zhen2023relightable}
Zhen Xu, Sida Peng, Chen Geng, Linzhan Mou, Zihan Yan, Jiaming Sun, Hujun Bao, and Xiaowei Zhou.
\newblock Relightable and animatable neural avatar from sparse-view video.
\newblock In \emph{CVPR}, 2024{\natexlab{b}}.

\bibitem[Yan et~al.(2024)Yan, Lin, Zhou, Wang, Sun, Zhan, Lang, Zhou, and Peng]{yan2024street}
Yunzhi Yan, Haotong Lin, Chenxu Zhou, Weijie Wang, Haiyang Sun, Kun Zhan, Xianpeng Lang, Xiaowei Zhou, and Sida Peng.
\newblock Street gaussians for modeling dynamic urban scenes.
\newblock In \emph{ECCV}, 2024.

\bibitem[Yang et~al.(2024{\natexlab{a}})Yang, Kang, Huang, Xu, Feng, and Zhao]{depthanything}
Lihe Yang, Bingyi Kang, Zilong Huang, Xiaogang Xu, Jiashi Feng, and Hengshuang Zhao.
\newblock {Depth Anything}: Unleashing the power of large-scale unlabeled data.
\newblock In \emph{CVPR}, 2024{\natexlab{a}}.

\bibitem[Yang et~al.(2024{\natexlab{b}})Yang, Gao, Zhou, Jiao, Zhang, and Jin]{yang2023deformable3dgs}
Ziyi Yang, Xinyu Gao, Wen Zhou, Shaohui Jiao, Yuqing Zhang, and Xiaogang Jin.
\newblock Deformable 3d gaussians for high-fidelity monocular dynamic scene reconstruction.
\newblock In \emph{CVPR}, 2024{\natexlab{b}}.

\bibitem[Yang et~al.(2024{\natexlab{c}})Yang, Yang, Pan, and Zhang]{yang2023gs4d}
Zeyu Yang, Hongye Yang, Zijie Pan, and Li Zhang.
\newblock Real-time photorealistic dynamic scene representation and rendering with 4d gaussian splatting.
\newblock In \emph{ICLR}, 2024{\natexlab{c}}.

\bibitem[Yu et~al.(2021{\natexlab{a}})Yu, Ye, Tancik, and Kanazawa]{yu2021pixelnerf}
Alex Yu, Vickie Ye, Matthew Tancik, and Angjoo Kanazawa.
\newblock {pixelNeRF}: Neural radiance fields from one or few images.
\newblock In \emph{CVPR}, 2021{\natexlab{a}}.

\bibitem[Yu et~al.(2021{\natexlab{b}})Yu, Zheng, Guo, Liu, Dai, and Liu]{tao2021function4d}
Tao Yu, Zerong Zheng, Kaiwen Guo, Pengpeng Liu, Qionghai Dai, and Yebin Liu.
\newblock {Function4D}: Real-time human volumetric capture from very sparse consumer rgbd sensors.
\newblock In \emph{CVPR}, 2021{\natexlab{b}}.

\bibitem[Zhang et~al.(2018)Zhang, Isola, Efros, Shechtman, and Wang]{Zhang_2018_CVPR}
Richard Zhang, Phillip Isola, Alexei~A. Efros, Eli Shechtman, and Oliver Wang.
\newblock The unreasonable effectiveness of deep features as a perceptual metric.
\newblock In \emph{CVPR}, 2018.

\bibitem[Zhao et~al.(2022)Zhao, Yang, Zhang, Lin, Zhang, Yu, and Xu]{Zhao_2022_CVPR}
Fuqiang Zhao, Wei Yang, Jiakai Zhang, Pei Lin, Yingliang Zhang, Jingyi Yu, and Lan Xu.
\newblock {HumanNeRF}: Efficiently generated human radiance field from sparse inputs.
\newblock In \emph{CVPR}, 2022.

\bibitem[Zheng et~al.(2024)Zheng, Zhou, Shao, Liu, Zhang, Nie, and Liu]{zheng2024gpsgaussian}
Shunyuan Zheng, Boyao Zhou, Ruizhi Shao, Boning Liu, Shengping Zhang, Liqiang Nie, and Yebin Liu.
\newblock {GPS-Gaussian}: Generalizable pixel-wise 3d gaussian splatting for real-time human novel view synthesis.
\newblock In \emph{CVPR}, 2024.

\bibitem[Zheng et~al.(2023)Zheng, Yifan, Wetzstein, Black, and Hilliges]{zheng2023pointavatar}
Yufeng Zheng, Wang Yifan, Gordon Wetzstein, Michael~J Black, and Otmar Hilliges.
\newblock {Pointavatar}: Deformable point-based head avatars from videos.
\newblock In \emph{CVPR}, 2023.

\bibitem[Zheng et~al.(2022)Zheng, Yu, Liu, and Dai]{zheng2020pamir}
Zerong Zheng, Tao Yu, Yebin Liu, and Qionghai Dai.
\newblock {PaMIR}: Parametric model-conditioned implicit representation for image-based human reconstruction.
\newblock \emph{IEEE Transactions on Pattern Analysis and Machine Intelligence}, 44\penalty0 (6):\penalty0 3170--3184, 2022.

\bibitem[Zhou et~al.(2024{\natexlab{a}})Zhou, Shao, Xu, Bai, Qiu, Liu, Wang, Geiger, and Liao]{Zhou_2024_CVPR}
Hongyu Zhou, Jiahao Shao, Lu Xu, Dongfeng Bai, Weichao Qiu, Bingbing Liu, Yue Wang, Andreas Geiger, and Yiyi Liao.
\newblock {HUGS}: Holistic urban 3d scene understanding via gaussian splatting.
\newblock In \emph{CVPR}, 2024{\natexlab{a}}.

\bibitem[Zhou et~al.(2024{\natexlab{b}})Zhou, Lin, Shan, Wang, Sun, and Yang]{zhou2024drivinggaussian}
Xiaoyu Zhou, Zhiwei Lin, Xiaojun Shan, Yongtao Wang, Deqing Sun, and Ming-Hsuan Yang.
\newblock {Drivinggaussian}: Composite gaussian splatting for surrounding dynamic autonomous driving scenes.
\newblock In \emph{CVPR}, 2024{\natexlab{b}}.

\end{thebibliography}
